\documentclass[11pt]{article}
\usepackage[final]{acl}

\usepackage{times}
\usepackage{latexsym}
\usepackage[T1]{fontenc}
\usepackage[utf8]{inputenc}
\usepackage{microtype}
\usepackage{inconsolata}
\usepackage{graphicx}

\usepackage{amsmath,amssymb}
\usepackage{booktabs}
\usepackage{tabularx}
\usepackage{array}
\usepackage{multirow}
\usepackage{pifont}
\usepackage{placeins}

\newcommand{\cmark}{\ding{51}}
\newcommand{\xmark}{\ding{55}}
\newcommand{\pmark}{$\circ$}

\newcommand{\orbit}{\textsc{Orbit}}

\title{\orbit: An Evidence-Grounded Benchmark for Evaluating LLMs\\on IT Operations Reasoning}

\author{Sathwik Narkedimilli$^{1}$, Rakshit Verma$^{2}$, Rahul Soni$^{2}$, Ankit Kushwaha$^{2}$, \\
  \textbf{Balaji Ganesan$^{3}$, Kaustabha Ray$^{4}$, Sunil Saumya$^{2}$, Pratibha Moogi$^{3}$} \\
  $^{1}$National University of Singapore, Singapore \\
  $^{2}$Indian Institute of Information Technology Dharwad, India \\
  $^{3}$IBM Software, India \quad
  $^{4}$Indian Association for the Cultivation of Science, India \\
  \small{\texttt{sathwik.narkedimilli@ieee.org}, \texttt{\{23bcs105, 23bcs104, 23BDS008, sunil.saumya\}@iiitdwd.ac.in}} \\
  \small{\texttt{bganesa1@in.ibm.com},
\small{\texttt{smcskr@iacs.res.in}}}, \texttt{PRATIBHA.Moogi@ibm.com}}

\begin{document}
\maketitle

\begin{abstract}
LLMs now support IT operations (ITOps), but benchmarks rarely show why a model fails. We present \textbf{ORBIT}, a benchmark of ITOps questions in three formats, namely multiple-choice, open-ended, and troubleshooting, with each question accompanied by a reference answer and a recorded source. An audit found template defects in about a tenth of the generated items, so we drop them and rebuild the troubleshooting format from real practitioner threads with accepted answers. Under one independent judge, five LLMs answer most multiple-choice items correctly, between a third and just over half of the open-ended items, and a minority of the troubleshooting items, even under a more lenient root-cause rubric. When models judged their own answers, they accepted many answers to defective items, so we use one independent judge for every model. Word overlap with the question finds the correct option for more than half of the multiple-choice items. ORBIT is a diagnostic aid for selecting and supervising assistants rather than a deployment test.
\end{abstract}

\section{Introduction}
\label{sec:intro}

Information Technology Operations (ITOps) keeps an organization's software running. It finds the causes of failures and restores service \citep{samanta2024efficient, sarma2026aiops, samanta2023insightssumm, deng2023ibm}. A typical setup runs containerized microservices under an orchestrator in the cloud. Application Performance Monitoring (APM) tools such as Datadog, Dynatrace, and SigNoz collect logs, traces, and metrics \citep{chandrachood2023optimizing}.

When service degrades, operators search this data for the cause and a fix. This is Site Reliability Engineering (SRE) work \citep{chigurupati2025site, panda2025benefits}. AI for IT Operations (AIOps) applies machine learning, and now LLMs and agents, to detect anomalies, find root causes, and suggest fixes \citep{rana2026aiops, malik2026edge, chen2025aiopslab, drosos2024your}.

Fixing faults is hard because modern infrastructure is tightly coupled and faults spread. One error in a Terraform template \citep{brikman2022terraform} can disturb scheduling on a Kubernetes cluster \citep{poniszewska2021kubernetes} and show up only as higher latency in an unrelated service. Solving it means linking deployment history, scaling rules, metrics, and documentation into one causal chain, not recalling a single fact.

Current evaluation does not measure this well \citep{pahune2025transitioning, zhou2024haicosystem}, and existing ITOps benchmarks target other skills \citep{sarma2027ai}. SWE-bench \citep{jimenez2024swe} tests code changes, ITBench \citep{jhaitbench} tests agents in live IT environments, and OpsEval \citep{liu2025opseval} tests broad operational knowledge. Their items do not link answers to source passages or reasoning steps, so a wrong answer cannot be traced to a missing fact or a broken step.

Building such a benchmark is hard for three reasons. (i) Production ITOps data is confidential, so realistic scenarios cannot be mined from incident logs. (ii) Operational questions easily collapse into keyword lookups. ORBIT only partly avoids this. By its own labels 62.6\% of items are lookups, and word overlap with the question finds 58\% of multiple-choice answers (\S\ref{sec:shortcut}). (iii) One accuracy number hides which abilities a model lacks.

We address these challenges with ORBIT. \emph{Objective} means that every item has a fixed reference answer. Multiple-choice items are scored by exact match. Open-ended and troubleshooting answers are compared with the reference by an LLM judge, so those scores are less objective (\S\ref{sec:experiments}). A human-in-the-loop pipeline builds items from operational documentation and practitioner discussions and records a source for 88\% of generated items. Only 798 items (25\%) cite public pages, since most sources sit in an unreleased internal repository.

Consider a multiple-choice item from ORBIT (inst-101, Appendix~\ref{app:examples}). PHP-FPM requests hang for 20 to 60 seconds after the page is generated. A keyword match on ``hang'' suggests slow code or a timeout. The recorded discussion shows a different cause, namely resource contention, while the Instana agent sent metrics over TCP. A model must connect trace data, packet captures, and monitoring behavior. The reference trace records these steps, so an incorrect answer can be identified.

ORBIT has three reasoning tiers, 62.6\% factual or schema retrieval, 10.6\% semantic interpretation, and 26.8\% abductive root-cause, causal multi-hop, or procedural reasoning. The evaluated diagnostic cohort holds 362 multiple-choice items (Figure~\ref{fig:taxonomy}).

\paragraph{Intended use.}
ORBIT helps teams compare LLM assistants for SRE before deployment, using per-format, per-type, and diagnostic scores to decide where humans must check. We have not yet used ORBIT to select a production model or compare its rankings with engineers' judgments on real incidents. The results thus support careful selection but do not prove fitness for production.

\paragraph{Lessons for teams.} We evaluate five LLMs closed-book on 3,019 items and draw four lessons. Multiple-choice accuracy overstates what a model can do in free text, since every model scores far lower on open-ended and troubleshooting items, and the ranking changes with the format (\S\ref{sec:experiments}). Difficulty labels predict failure, since hard multiple-choice items are about 30 points harder than standard ones in the same format. Diagnosis of a real fault is the weakest skill, so engineers must check every diagnosis. Comparing models needs one independent judge, since self-judging reorders them. Our experiments do not yet use the recorded evidence or reasoning traces; we leave that to future work.

Our contributions are as follows.
\begin{itemize}
\setlength\itemsep{0pt}
\item \textbf{A construction method} that records the evidence behind answers, rebuilds troubleshooting items from practitioner threads with a cross-vendor audit, and reports lessons on judging and item quality (\S\ref{sec:construction}).
\item \textbf{The ORBIT benchmark} has 3,019 evaluation items across three formats and several reasoning tiers, each with a reference answer, and 96\% include a recorded source.
\item \textbf{A closed-book study} of five LLMs with one independent judge, corrected significance tests, effect sizes, and the smallest model difference ORBIT can detect (\S\ref{sec:experiments}).
\item \textbf{An anonymized release} of the code and data.\footnote{\url{https://anonymous.4open.science/r/orbit-BC6B}}
\end{itemize}

\section{Related Work}
\label{sec:related}

This section covers benchmarks, while Appendix~\ref{app:related} reviews AIOps and LLM-for-ITOps work.

\paragraph{Benchmarks.}
At SWE-bench's release, the best model with BM25 retrieval, Claude~2, resolved 1.96\% of 2,294 GitHub issues \citep{jimenez2024swe}. In ITBench, GPT-4o scored 0\% on FinOps and 13.8\% on SRE tasks due to non-deterministic execution and partial observability \citep{jhaitbench}. HotpotQA \citep{yang2018hotpotqa} pairs multi-hop questions with encyclopedic evidence.

OpsEval \citep{liu2025opseval} is ORBIT's closest neighbor. Both ask operational questions in multiple-choice and open-ended forms. OpsEval items carry an answer but no link to a source passage and no reasoning trace. ORBIT items add a recorded source, a reference trace, a reasoning-type label, and a difficulty label. OpsEval also notes privacy limits on operational data, which ORBIT meets by citing internal sources by path. Owl-Bench \citep{guo2024owl}, released with OWL, likewise tests broad IT-operations knowledge.

\paragraph{Telemetry and agent benchmarks.}
OpenRCA \citep{xu2025openrca} and RCAEval \citep{pham2025rcaeval} score root-cause localization over telemetry. AIOpsLab \citep{chen2025aiopslab}, SREGym \citep{clark2026sregym}, and Cloud-OpsBench \citep{wang2026cloudopsbench} test agents in live or snapshotted clusters, and LogEval \citep{cui2024logeval} covers log analysis. None of them link answers to documented evidence. Closest to our troubleshooting items, \citet{ahmed2023recommending} have incident owners rate LLM root-cause and mitigation suggestions for real cloud incidents.

\paragraph{Requirements and gap.}
We derive four requirements (Appendix~\ref{app:requirements}), namely operational grounding, multi-entity reasoning, evidence traceability, and technology diversity. Table~\ref{tab:benchmark_comparison} compares benchmarks on the first three plus reasoning annotations. ITBench, AIOpsLab, SREGym, and Cloud-OpsBench are execution-centric, OpenRCA and RCAEval localize faults in telemetry, OpsEval is knowledge-oriented, and SWE-bench scores code patches. The criteria are our own, judged from each benchmark's published description, so the table shows how benchmarks differ rather than which is better. ORBIT is partial on multi-entity reasoning because 26.8\% of its items are reasoning-heavy, and on evidence traceability because 88\% have a recorded source but only 25\% a public one.

ORBIT combines operational sources, recorded evidence, reference traces, several formats, and difficulty labels. We do not claim that other benchmarks lack reasoning. ORBIT complements execution-centric benchmarks. Assistants built into observability platforms cannot be compared, as their models and data are private (Appendix~\ref{app:discovery}).

\begin{table}[t]
\centering
\small
\setlength{\tabcolsep}{4pt}
\resizebox{\columnwidth}{!}{%
\begin{tabular}{lccccc}
\toprule
\textbf{Benchmark} & \textbf{Set.} & \textbf{OG} & \textbf{ME} & \textbf{ET} & \textbf{RA} \\
\midrule
SWE-bench \citep{jimenez2024swe} & E & \xmark & \cmark & \cmark & \xmark \\
ITBench \citep{jhaitbench} & E & \cmark & \cmark & \xmark & \xmark \\
AIOpsLab \citep{chen2025aiopslab} & E & \cmark & \cmark & \xmark & \xmark \\
SREGym \citep{clark2026sregym} & E & \cmark & \cmark & \xmark & \xmark \\
Cloud-OpsBench \citep{wang2026cloudopsbench} & E & \cmark & \cmark & \xmark & \xmark \\
OpenRCA \citep{xu2025openrca} & T & \cmark & \cmark & \xmark & \xmark \\
RCAEval \citep{pham2025rcaeval} & T & \cmark & \pmark & \xmark & \xmark \\
LogEval \citep{cui2024logeval} & T & \cmark & \xmark & \xmark & \xmark \\
OpsEval \citep{liu2025opseval} & Q & \cmark & \pmark & \xmark & \xmark \\
GraphQLRestBench \citep{saha2024sequential} & Q & \xmark & \cmark & \cmark & \xmark \\
\midrule
\textbf{ORBIT (ours)} & Q & \cmark & \pmark & \pmark & \cmark \\
\bottomrule
\end{tabular}}
\par\smallskip{\scriptsize Set.\ is the setting, where E is execution, T is telemetry or log analysis, and Q is question answering. OG (operational grounding) means items come from operational material, ME (multi-entity reasoning) means items need several pieces of evidence, ET (evidence traceability) means answers link to source passages, and RA (reasoning annotations) means reference traces are released per item. \cmark\ is yes, \xmark\ ,is no, and \pmark\ is partial.\par}
\caption{How ORBIT compares with other ITOps benchmarks on operational grounding, multi-entity reasoning, evidence traceability, and reasoning annotations.}
\label{tab:benchmark_comparison}
\end{table}

\section{Building ORBIT}
\label{sec:construction}

\begin{figure*}[t]
  \centering
  \includegraphics[width=0.95\textwidth]{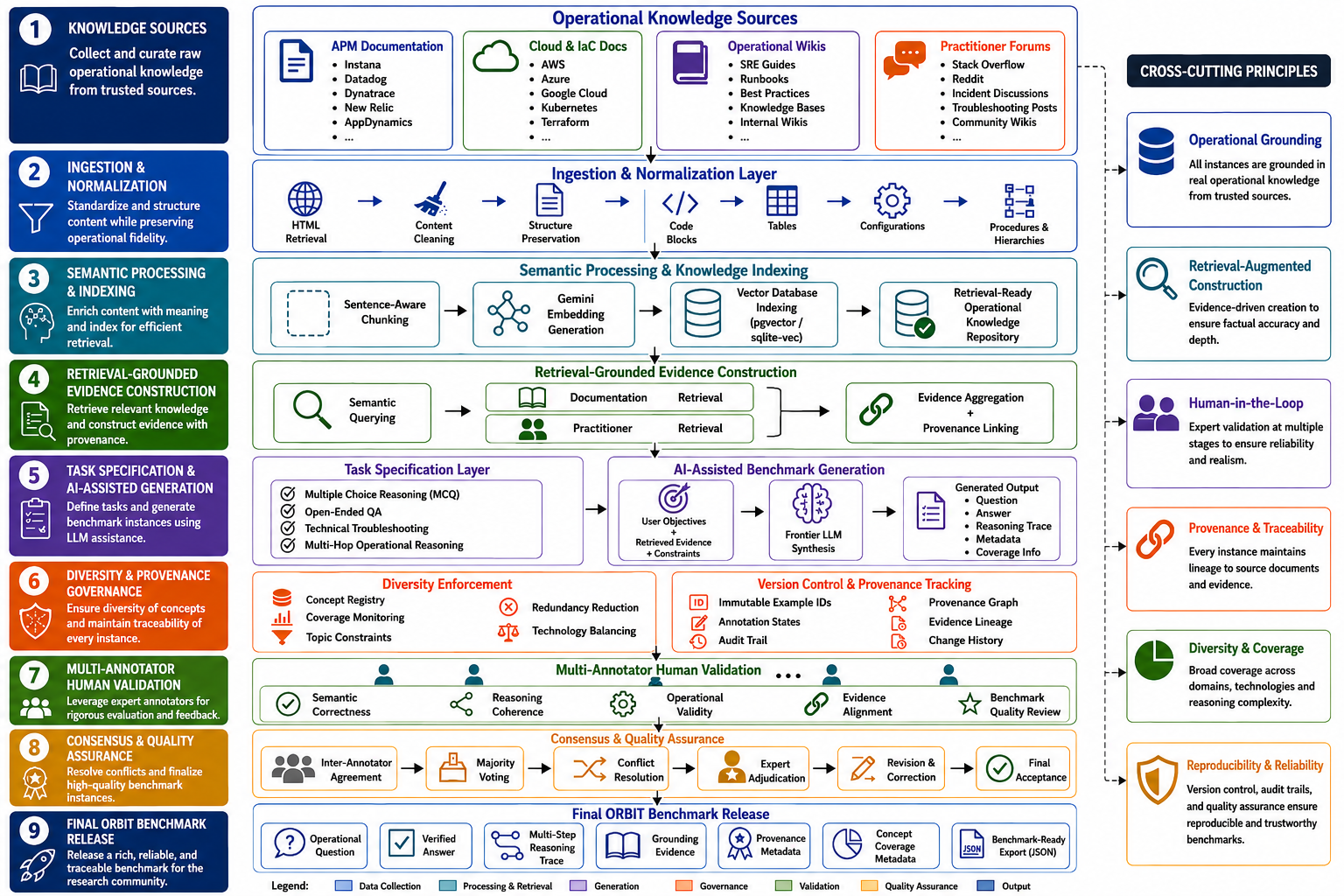}
  \caption{The ORBIT pipeline, from collecting and indexing sources, through evidence retrieval and LLM-assisted item generation, to validation and consensus checks before release.}
  \label{fig:pipeline}
\end{figure*}

A human-in-the-loop pipeline (Figure~\ref{fig:pipeline}, Appendix~\ref{app:construction}) has four stages, (A) assembling a knowledge base, (B) generating evidence-grounded items, (C) labeling difficulty, reasoning type, and cohort, and (D) expert validation and cleaning. Step (E) then rebuilds the troubleshooting format.

\paragraph{(A) Knowledge base.}
ORBIT draws on seven source collections in four groups. These include APM documentation for Instana, Dynatrace, and Datadog, along with cloud and infrastructure documentation, operational wikis and runbooks (including an internal repository), and practitioner forums such as Stack Overflow \citep{barua2014developers}. Of the 2,803 items with a recorded source, 2,005 cite the internal repository, and 798 cite public pages (794 Stack Overflow threads and 4 Wikipedia pages). No released item is grounded in Reddit.

Coverage is uneven. Instana grounds about 1,360 items, seven other IT products about 570, and AWS, Azure, and GCP only about 25.

\paragraph{(B) Retrieval-grounded generation.}
For a target topic, semantic queries retrieve passages from documentation, which holds system facts, and from practitioner threads, which hold troubleshooting patterns. Gemini Flash (\texttt{gemini-3-flash-preview} by default, or \texttt{gemini-2.5-flash}) receives this evidence and the task rules and drafts an item whose answer follows from it. The item stores its source identifiers. Retrieval settings are in Appendix~\ref{app:construction}. A concept registry, coverage tracking, and topic rules favor rare concepts and limit near-duplicates.

\paragraph{(C) Difficulty and reasoning labels.}
Each item gets a difficulty score that combines a rule-based score with an LLM review of reasoning, evidence, and troubleshooting. Items above a fixed threshold are labeled \emph{hard} (740, 23.3\%) and the rest standard (2,438, 76.7\%). No item is perturbed or written to fool a model. Because the score rewards troubleshooting complexity, hard items are mostly generative (478 open-ended, all 133 troubleshooting, and 129 multiple-choice). An LLM also assigns a primary reasoning type and a failure-mode cohort, which are not checked against human labels.

\paragraph{(D) Expert validation and cleaning.}
Three of the authors, all with software engineering and ITOps experience, reviewed every candidate for correctness, reasoning, operational validity, and answer-evidence fit. They accepted, revised, or rejected it, with a majority vote for disputes and expert adjudication for harder cases. Cleaning then removed duplicates and repaired broken items while keeping each validated answer fixed (Appendix~\ref{app:construction}). During cleaning, Llama 3.3 70B rewrote 470 open-ended reference answers and 113 sets of distractors. These 583 items were not re-reviewed, and the other 2,595 were not rewritten after validation.

Llama 3.3 70B, also an evaluated model, scores lowest on open-ended items under the independent judge (\S\ref{sec:experiments}), so its rewrites do not appear to favor it.

\paragraph{(E) Troubleshooting from practitioner threads.}
Every generated troubleshooting item was a template (\S\ref{sec:benchmark}), so we rebuilt this format from Stack Overflow, Server Fault, and DevOps Stack Exchange threads with accepted answers. We sampled 438 threads labeled abductive root cause that no other item cites. Claude Opus 5.5 rewrote each question as an incident report that keeps evidence such as error messages, versions, and configuration but not the cause. It wrote a reference root cause, fix, and key points based only on the accepted answer. GPT-5.5, from another vendor, checked each item against its thread. An item was kept only if its reference was faithful, its scenario hid the cause, the cause could be reached from the scenario, and there was one primary cause. 208 items passed. Most rejections revealed the cause (Appendix~\ref{app:ts}). Experts have not yet reviewed the kept items.

\paragraph{Lessons from construction.}
First, items built from documentation are almost all lookups. Of the 2,005 items grounded in the internal repository, 1,953 are retrieval items, while reasoning-heavy items come almost entirely from Stack Overflow threads. Second, generated items need an independent answer check because generated numerical items often have arithmetic errors. Third, the judge must be strict. A lenient pilot judge accepted 90--97\% of open-ended answers for every model, roughly 40 to 50 points above a strict prompt. Fourth, a model must not judge its own answers, since LLM judges favor their own outputs \citep{panickssery2024llm, zheng2023judging}. For 287 items whose references could not be matched, self-judges accepted 7 to 75\% of answers, and the independent judge accepted 3 to 5\% (Appendix~\ref{app:allitems}).

\section{The ORBIT Benchmark}
\label{sec:benchmark}

\paragraph{Composition.}
Of the 3,178 unique items, 79.1\% are multiple-choice, 16.7\% open-ended, and 4.2\% troubleshooting. Questions average $14.28 \pm 5.56$ whitespace tokens (13.6 for multiple-choice, 14.9 for open-ended, and 25.6 for troubleshooting) and name $3.19 \pm 1.72$ technical entities such as commands or error codes (Table~\ref{tab:orbit_feature_statistics}). LLM-assigned reasoning types form three tiers, namely factual or schema retrieval (1,989, 62.6\%), semantic interpretation (338, 10.6\%), and reasoning-heavy (851, 26.8\%), led by abductive root cause (610).

\paragraph{Cohorts.}
Failure-mode cohorts group items by the reasoning they need across formats. In the generated corpus, the diagnostic cohort has 680 items (362 multiple-choice, 187 open-ended, and 131 troubleshooting), and only the 362 multiple-choice ones remain after the audit. The multi-hop causal cohort has 253 items, mostly open-ended. Appendix~\ref{app:examples} gives examples, including a full troubleshooting item. Reference traces may include more steps than an item needs, so tier labels describe the question, not its trace length.

\paragraph{Provenance coverage.}
Of the 3,178 items, 2,803 (88.2\%) have at least one recorded source, and 299 (9.4\%) have two or more. 798 items cite public pages such as Stack Overflow threads. The other 2,005 grounded items cite an internal documentation repository by path and pinned commit, which is not redistributed. A release audit confirmed that 2,001 of these links resolve to valid pages, while 4 record only a document title. A valid link does not prove that the page supports the answer, and we have not yet checked support manually.

The remaining 375 items (11.8\%) have no recorded source, including 175 short variants generated from templates. Users can filter on the grounding field to keep only grounded items.

\paragraph{Evaluation set.}
An audit of every generative item found 287 with no matchable reference. These are 175 template variants (140 open-ended and 35 troubleshooting) with a fixed answer phrase, 58 troubleshooting items answered by a bare option letter, and 54 open-ended items whose reference repeats a forum question. Another 40 open-ended and 40 troubleshooting items are ungrounded templates. We exclude all 367 and add the 208 rebuilt troubleshooting items. The evaluation set has 3,019 items (2,513 multiple-choice, 298 open-ended, and 208 troubleshooting), and 96\% have a recorded source. Appendix~\ref{app:allitems} covers all generated items.

\paragraph{Grader study and topical coverage.}For
Human consensus and a three-LLM ensemble agree ($\kappa=0.80$) on 5-point grades of 100 responses (Appendix~\ref{app:iaa}), and 75.05\% of items match a Stack Overflow thread by topic (Appendix~\ref{app:external}). Neither validates ORBIT's labels or the judge behind Tables~\ref{tab:main_results}--\ref{tab:failure_modes}. Many items come from such threads, so topic match shows overlap, not correctness.

\subsection{Surface Cues and Accuracy}
\label{sec:shortcut}

A multiple-choice item tests knowledge only if its answer cannot be guessed from surface cues \citep{gururangan2018annotation, balepur2024artifacts}. Probes adapted from SIMBA-UQ \citep{simba_uq} see only the four options and score them by word overlap. The best reaches top-choice accuracy of 0.33 (AUROC 0.59), close to always choosing B or the longest option (0.31). These option-only probes never see the question (Appendix~\ref{app:shortcut}).

A probe that also sees the question finds a strong cue. Choosing the option with the most word overlap with the question is right on 57.9\% of items. It reaches 60.5\% on items from the internal repository, 33.1\% on public items, and 94.6\% on the 120 ungrounded items. In 102 items, only the correct option uses code formatting. An options-only run that hides the question scores 30.9\% with Llama 3.3 70B and 32.7\% with Mistral Medium, near the 25\% chance rate. The options alone, therefore, rarely reveal the answer, except for the 120 ungrounded items, which both models answer without the question (Appendix~\ref{app:shortcut}); these items test nothing.

\section{Experiments}
\label{sec:experiments}

\paragraph{Setup.}
We evaluate Llama 3.3 70B Instruct, Mistral Small 3.1 (24B), Llama 4 Maverick (17B-128E), Granite 4.0 H Small, and Mistral Medium (2505), all served through watsonx.ai. Four have open weights, and one is API-only. They cover dense, mixture-of-experts, and hybrid state-space designs. No frontier or reasoning model is included, so the accuracy ceiling on ORBIT is unknown.

\paragraph{Closed-book setting.}
Models see the question and, for multiple-choice items, the options, but no evidence. This measures what a model brings to an incident without retrieval, as when an engineer asks directly. Production assistants retrieve information from runbooks and documentation, and ORBIT stores that evidence, making open-book evaluation more realistic. We have not run it (Appendix~\ref{app:contamination}).

\paragraph{Scoring.}
Answers use greedy decoding via the watsonx.ai chat interface, and multiple-choice answers are scored by exact match. Granite 4.0 H Small's multiple-choice answers come from an earlier run through the text interface, as its chat endpoint queued requests for minutes. On the 216 items that both runs scored, each got 186 right. Claude Sonnet 5.5 judges open-ended and troubleshooting answers against the reference answer and explanation at temperature 0 with a strict prompt (Appendix~\ref{app:protocol}). One judge puts all models on a common scale and is not itself evaluated, but its agreement with humans is not yet measured. A more lenient second judge from another vendor agrees with it on 75.7\% of open-ended and 89.6\% of troubleshooting verdicts (Appendix~\ref{app:secondjudge}).

\paragraph{Statistical analysis.}
We report 95\% Wilson intervals from counts and use two-proportion $z$-tests with Cohen's $h$ as effect size (0.2 small, 0.5 medium, 0.8 large) and Holm's correction within each family. Format, hard-item, and cohort comparisons use disjoint items. For shared items, the unpaired test is more conservative than McNemar's test when errors correlate positively (Appendix~\ref{app:stats}).

\begin{table}[t]
\centering
\scriptsize
\setlength{\tabcolsep}{2.5pt}
\resizebox{\columnwidth}{!}{%
\begin{tabular}{lcccc}
\toprule
\textbf{Model} & \textbf{Overall} & \textbf{MCQ} & \textbf{OE} & \textbf{TS} \\
& ($n{=}3{,}019$) & ($n{=}2{,}513$) & ($n{=}298$) & ($n{=}208$) \\
\midrule
Llama 3.3 70B & 78.0 (76.5--79.5) & 88.1 (86.8--89.3) & 35.6 (30.4--41.2) & 16.8 (12.4--22.5) \\
Mistral Small 3.1 & 79.2 (77.7--80.6) & 88.4 (87.1--89.6) & 47.7 (42.0--53.3) & 13.5 (9.5--18.8) \\
Llama 4 Maverick & 77.8 (76.3--79.2) & 85.8 (84.4--87.1) & 45.3 (39.7--51.0) & \textbf{26.9} (21.4--33.3) \\
Granite 4.0 H & 74.1 (72.5--75.6) & 81.5 (79.9--82.9) & \textbf{54.7} (49.0--60.3) & 12.5 (8.7--17.7) \\
Mistral Medium & \textbf{80.3} (78.8--81.6) & \textbf{89.0} (87.7--90.2) & 50.7 (45.0--56.3) & 16.8 (12.4--22.5) \\
\bottomrule
\end{tabular}}
\par\smallskip{\scriptsize MCQ is multiple-choice, OE is open-ended, and TS is troubleshooting. Best results are in bold.\par}
\caption{Zero-shot accuracy (\%) of each model on the evaluation set, with Wilson confidence intervals.}
\label{tab:main_results}
\end{table}

\paragraph{Multiple-choice accuracy overstates free-text skill.}
Table~\ref{tab:main_results} shows the same order across all models. Multiple-choice accuracy (81.5--89.0\%) far exceeds open-ended accuracy (35.6--54.7\%), which exceeds troubleshooting accuracy (12.5--26.9\%). A root-cause rubric, added after we found that the strict prompt fails right diagnoses that omit a side detail, gives 18.3--40.9\% (Appendix~\ref{app:ts}). Under a more lenient second judge, the drop to open-ended accuracy shrinks to 6.6--24.0 points, while troubleshooting accuracy stays far lower (Appendix~\ref{app:secondjudge}). Rankings change with the format. Granite 4.0 H Small is last on multiple-choice and troubleshooting items but first on open-ended items, while Llama 3.3 70B is third on multiple-choice items and last of the five on open-ended items.

The drop from multiple-choice to open-ended accuracy is significant for every model ($p_{\text{Holm}}<10^{-24}$, $h=0.59$--$1.16$), and so is the drop from multiple-choice to troubleshooting accuracy ($p_{\text{Holm}}<10^{-95}$, $h=1.28$--$1.69$). Troubleshooting accuracy is significantly below open-ended accuracy for all five models. Pooled over all items, four of ten model pairs differ significantly, each involving Granite 4.0 H Small, and every pairwise effect is negligible ($|h|<0.2$). On open-ended items, three pairs differ, each involving Llama 3.3 70B; for troubleshooting items, two pairs differ, both involving Llama 4 Maverick (Appendix~\ref{app:stats}).

These differences fit a gap between recognizing and generating answers, but do not isolate it. Multiple-choice and generative items are different item sets with different scorers. The gap therefore also reflects item difficulty, answer specificity, and judge behavior. A controlled test would strip the options from the same multiple-choice items and score the open answers with one judge. Since question-option overlap alone finds 58\% of multiple-choice answers (\S\ref{sec:shortcut}), part of the recognition advantage may be cue use. Troubleshooting items also come from real threads and are longer.

\begin{table}[t]
\centering
\scriptsize
\setlength{\tabcolsep}{3pt}
\resizebox{\columnwidth}{!}{%
\begin{tabular}{lccc}
\toprule
\textbf{Model} & \textbf{Standard} ($n{=}2{,}384$) & \textbf{Hard} ($n{=}129$) & \textbf{$\Delta$} \\
\midrule
Llama 3.3 70B & 89.7 (88.4--90.8) & 59.7 (51.1--67.8) & 30.0 \\
Mistral Small 3.1 & 90.1 (88.8--91.2) & 56.6 (48.0--64.8) & 33.5 \\
Llama 4 Maverick & 87.3 (85.9--88.6) & 58.1 (49.5--66.3) & \textbf{29.2} \\
Granite 4.0 H & 83.1 (81.5--84.5) & 51.2 (42.6--59.6) & 31.9 \\
Mistral Medium & 90.6 (89.3--91.7) & 60.5 (51.8--68.5) & 30.1 \\
\bottomrule
\end{tabular}}
\par\smallskip{\scriptsize$\Delta$ is standard minus hard accuracy, and the smallest gap is in bold.\par}
\caption{Multiple-choice accuracy (\%) on standard and hard items, with Wilson confidence intervals.}
\label{tab:adversarial}
\end{table}

\paragraph{Difficulty labels predict failure.}
Table~\ref{tab:adversarial} compares hard and standard items within the multiple-choice format, so both groups are scored by exact match. Hard items are 29.2 to 33.5 points harder for every model ($p_{\text{Holm}}<10^{-19}$, $h=0.68$--$0.80$), with hard-item accuracy of 51.2--60.5\%. Open-ended items allow no such test, since 285 of the 298 are hard. Earlier few-shot results come from a self-judged pass that the released harness cannot reproduce, so they appear only in Appendix~\ref{app:oldpass}.

\begin{table}[t]
\centering
\scriptsize
\setlength{\tabcolsep}{2.5pt}
\resizebox{\columnwidth}{!}{%
\begin{tabular}{lcccc}
\toprule
& \multicolumn{2}{c}{\textbf{Multiple-choice}} & \multicolumn{2}{c}{\textbf{Generative}} \\
\cmidrule(lr){2-3}\cmidrule(lr){4-5}
\textbf{Model} & \textbf{Diagnostic} & \textbf{Other} & \textbf{OE multi-hop} & \textbf{TS} \\
& ($n{=}362$) & ($n{=}2{,}151$) & ($n{=}235$) & ($n{=}208$) \\
\midrule
Llama 3.3 70B & 59.1 & 93.0 & 41.3 & 16.8 \\
Mistral Small 3.1 & 60.8 & 93.0 & 54.5 & 13.5 \\
Llama 4 Maverick & 59.1 & 90.3 & 53.2 & 26.9 \\
Granite 4.0 H & 54.1 & 86.1 & 63.0 & 12.5 \\
Mistral Medium & 63.3 & 93.4 & 57.9 & 16.8 \\
\bottomrule
\end{tabular}}
\par\smallskip{\scriptsize Diagnostic is the diagnostic cohort of multiple-choice items, and Other is the remaining multiple-choice items. OE multi-hop is the multi-hop causal cohort of open-ended items, and TS is the troubleshooting set.\par}
\caption{Accuracy (\%) on diagnosis items and on other items within each format.}
\label{tab:failure_modes}
\end{table}

\paragraph{Diagnosis of real faults is the weakest skill.}
Table~\ref{tab:failure_modes} contrasts diagnosis with other reasoning within each format. Among multiple-choice items, the 362 diagnostic items are 30.1 to 33.9 points harder than the other 2,151 for every model ($p_{\text{Holm}}<10^{-45}$, $h=0.72$--$0.85$). Among generative items, troubleshooting accuracy is 24.4 to 50.5 points below that on the 235 multi-hop open-ended items, significantly for every model ($h=0.54$--$1.11$). The generative contrast compares items from different pipelines, so it bounds rather than isolates the effect of diagnosis on model accuracy.

\paragraph{Accuracy by reasoning type.}
Table~\ref{tab:by_type} (Appendix~\ref{app:stats}) splits the 2,513 multiple-choice items by reasoning tier. Retrieval items are easy (90.4--97.1\%), while reasoning-heavy (53.8--63.2\%) and semantic items (31.2--50.7\%) are much harder. Every contrast with retrieval is significant ($p_{\text{Holm}}<10^{-72}$, $h=0.87$--$1.33$). Retrieval makes up 79\% of multiple-choice items, so the multiple-choice score mostly measures documentation recall. Users who care about reasoning should report reasoning-heavy and troubleshooting items separately.

\paragraph{Smallest detectable difference.}
With 3,019 items per model and accuracy near 78\%, an unpaired test at $\alpha=0.05$ and 80\% power detects an overall difference of about 3.0 points. The matching figures are about 11.4 points on the 298 open-ended items and about 10.4 points on the 208 troubleshooting items. All significant overall model differences are negligible in size ($h<0.2$), while every multiple-choice to open-ended drop exceeds $h=0.5$. ORBIT separates task types far more reliably than it separates these five models.

\paragraph{Implications for practice.}
For teams choosing an ITOps assistant, the results suggest four rules. First, do not choose on multiple-choice accuracy alone. Open-ended accuracy is 27 to 53 points lower (7 to 24 under the second judge), and model order changes with format. Second, set a troubleshooting accuracy floor before diagnoses reach on-call engineers. Even under the root-cause rubric, no model diagnoses over 40.9\% of real faults, so engineers must check every diagnosis.

Third, fix one independent judge before comparing models. Self-judging moved Llama 3.3 70B from second to last on open-ended items and accepted up to 75\% of answers to defective items. Fourth, ignore overall gaps below about 3.0 points, the size of most overall differences here.

\section{Conclusion}
\label{sec:conclusion}

We presented ORBIT, an operationally grounded ITOps benchmark spanning documentation recall to multi-step diagnosis. After removing defective templates and rebuilding troubleshooting items from practitioner threads, nearly all items have traceable sources. An independent judge found that five LLMs perform well on multiple-choice questions but struggle with free-text fault diagnosis, even under lenient RCS. Hard and diagnostic multiple-choice items remain challenging. These findings favor human-supervised assistant selection over autonomous remediation. Future work includes expert validation, open-book evaluation, and frontier reasoning models.

\clearpage
\section*{Limitations}
\label{sec:limitations}

\paragraph{Closed-book construct.}
Models do not see the evidence an item was built from. Our results measure operational question answering on source-grounded items, not reasoning over supplied evidence, and a high score may partly reflect memorized knowledge. ORBIT's recorded evidence and reasoning traces set it apart from OpsEval, yet no experiment uses them. An open-book run on the 2,803 grounded items and a trace-based error analysis are the most important missing experiments.

\paragraph{Scoring.}
Generative answers are graded by one LLM judge with a binary prompt. The judge is independent of the evaluated models, but its agreement with human grading is unmeasured, a second judge agrees only moderately on open-ended answers ($\kappa=0.53$), and a binary verdict hides partial credit. The 5-point study in Appendix~\ref{app:iaa} uses other graders and cannot validate the judge. Claude models wrote and judged the troubleshooting items, which may favor answers phrased like their references, though no Claude model is evaluated. Troubleshooting scores depend on the rubric (Appendix~\ref{app:ts}). Temperature 0 does not guarantee identical outputs across provider versions, so reruns may differ slightly from the reported results.

\paragraph{Models, sampling, and power.}
We evaluate five mid-2025 models from one hosting platform, with no frontier or reasoning models. Granite 4.0 H Small's multiple-choice answers came from a different interface than the others'. The open-ended and troubleshooting sets detect only differences of about 11.4 and 10.4 points. Our tests are unpaired, and paired tests would be more powerful.

\paragraph{Labels.}
Hard labels come from a difficulty threshold and are unpaired. Only 129 multiple-choice items are hard, but nearly all clean open-ended ones are. Reasoning-type and cohort labels are LLM-assigned and unchecked by humans. The corpus is imbalanced, with 62.6\% lookups and almost no numerical or temporal reasoning. Some multiple-choice items ask what an entry in a documentation hierarchy is, which tests knowledge of one repository, not operations. We release per-type labels for filtering and reweighting the corpus.

\paragraph{Data quality.}
Three authors were validators, which can bias the validation. After validation, Llama 3.3 70B rewrote 583 items without re-review. We excluded 367 items with template defects from evaluation and will repair or remove them in the public release. The 208 troubleshooting items were checked by a second LLM, not by expert validators, and a 30-item expert review is pending.

\paragraph{Evidence and contamination.}
Only 798 items (25\%) cite public pages that outsiders can check. The internal repository is not fully private in content, since 266 of its cited pages are saved copies of public vendor documentation. Its 1,678 cited schema pages come from an internal wiki.

Accuracy by provenance is reported only for multiple-choice items (Appendix~\ref{app:shortcut}). Reasoning-heavy items come mostly from public Stack Overflow threads and may be memorized. The troubleshooting threads were posted between 2009 and 2022, before the evaluated models were trained, so the models may recall the accepted answers. Rewriting each question as a scenario reduces verbatim overlap but does not remove this risk. We have not measured how well rewriting and duplicate removal limit verbatim overlap. Correct answers are mildly skewed toward option B (30.8\% of items).

\paragraph{External validity.}
Topic coverage measures overlap with public posts, not real-world validity. We have not linked accuracy to outcomes like diagnosis time or fix success, and report no production use. A study with on-call engineers must show that rankings predict usefulness on real incidents.

\section*{Ethical Considerations}

\paragraph{Data licensing and release.}
ORBIT is released as questions, reference answers, reasoning traces, and source references, without copying third-party content. Public pages such as Stack Overflow threads and vendor documentation are linked. Internal documentation contains infrastructure identifiers and personal data, so it is cited by path and pinned commit and not released. Code is released under the MIT license and data under CC BY-SA 4.0. This matches the Stack Overflow content license and allows fixes, subsets, and commercial use that industry users need. The troubleshooting scenarios paraphrase Stack Exchange posts and link to each thread, as required by the license requires.

\paragraph{Personal data.}
Items built from internal documentation may repeat their identifiers. Before public release, we will screen all released questions, answers, traces, and personal names for host names, IP addresses, and e-mail addresses, and mask or remove any found. We will also check the terms of use of the cited vendor documentation.

\paragraph{Use of AI assistance.}
Following the ACL policy, we disclose all AI use. ORBIT items were written with retrieval-grounded LLM assistance. Three authors validated every generated item under the consensus protocol (\S\ref{sec:construction}), and Llama 3.3 70B then rewrote some answers and distractors during the cleaning process. LLMs also assigned difficulty and reasoning labels (\S\ref{sec:construction}). Claude Opus 5.5 drafted the troubleshooting items, GPT-5.5 audited them, and Claude Sonnet 5.5 judged generative answers (\S\ref{sec:experiments}). AI assistants helped edit the language, and the authors checked all resulting text and numbers.

\bibliography{ref}

@inproceedings{saha2024sequential,
  title={Sequential {API} Function Calling Using {GraphQL} Schema},
  author={Saha, Avirup and Mandal, Lakshmi and Ganesan, Balaji and Ghosh, Sambit and Sindhgatta, Renuka and Eberhardt, Carlos and Debrunner, Dan and Mehta, Sameep},
  booktitle={Proceedings of the 2024 Conference on Empirical Methods in Natural Language Processing},
  pages={19452--19458},
  year={2024}
}

@inproceedings{jhaitbench,
  title={{ITBench}: Evaluating {AI} Agents across Diverse Real-World {IT} Automation Tasks},
  author={Jha, Saurabh and Arora, Rohan R and Watanabe, Yuji and Yanagawa, Takumi and Chen, Yinfang and Clark, Jackson and Bhavya, Bhavya and Verma, Mudit and Kumar, Harshit and Kitahara, Hirokuni and others},
  booktitle={Proceedings of the 42nd International Conference on Machine Learning},
  series={Proceedings of Machine Learning Research},
  volume={267},
  pages={27134--27197},
  url={https://proceedings.mlr.press/v267/jha25a.html},
  year={2025}
}

@inproceedings{dang2019aiops,
  title={{AIOps}: Real-World Challenges and Research Innovations},
  author={Dang, Yingnong and Lin, Qingwei and Huang, Peng},
  booktitle={2019 IEEE/ACM 41st International Conference on Software Engineering: Companion Proceedings (ICSE-Companion)},
  pages={4--5},
  year={2019},
  organization={IEEE}
}

@inproceedings{liu2025opseval,
  title={{OpsEval}: A Comprehensive Benchmark Suite for Evaluating Large Language Models' Capability in {IT} Operations Domain},
  author={Liu, Yuhe and Pei, Changhua and Xu, Longlong and Chen, Bohan and Sun, Mingze and Zhang, Zhirui and Sun, Yongqian and Zhang, Shenglin and Wang, Kun and Zhang, Haiming and others},
  booktitle={Proceedings of the 33rd ACM International Conference on the Foundations of Software Engineering},
  pages={503--513},
  year={2025}
}

@inproceedings{shan2025rca,
  title={{RCA} Copilot: Transforming Network Data into Actionable Insights via Large Language Models},
  author={Shan, Alexander and Kaur, Jasleen and Singh, Rahul and Banka, Tarun and Yavatkar, Raj and Sridhar, T},
  booktitle={ICC 2025-IEEE International Conference on Communications},
  pages={1566--1571},
  year={2025},
  organization={IEEE}
}

@article{zheng2023survey,
  title={A Survey of Large Language Models for Code: Evolution, Benchmarking, and Future Trends},
  author={Zheng, Zibin and Ning, Kaiwen and Wang, Yanlin and Zhang, Jingwen and Zheng, Dewu and Ye, Mingxi and Chen, Jiachi},
  journal={arXiv preprint arXiv:2311.10372},
  year={2023}
}

@inproceedings{sacco2025intent,
  title={Intent-Based {Kubernetes} Configuration via {LLMs}: Current Trends and Open Challenges},
  author={Sacco, Alessio and Zilli, Cristian and Marchetto, Guido},
  booktitle={2025 IEEE 50th Conference on Local Computer Networks (LCN)},
  pages={1--7},
  year={2025},
  organization={IEEE}
}

@inproceedings{simba_uq,
    title = "{SIMBA} {UQ}: Similarity-Based Aggregation for Uncertainty Quantification in Large Language Models",
    author = "Bhattacharjya, Debarun  and
      Ganesan, Balaji  and
      Lee, Junkyu  and
      Marinescu, Radu  and
      Mirylenka, Katya  and
      Glass, Michael  and
      Shou, Xiao",
    editor = "Christodoulopoulos, Christos  and
      Chakraborty, Tanmoy  and
      Rose, Carolyn  and
      Peng, Violet",
    booktitle = "Findings of the Association for Computational Linguistics: EMNLP 2025",
    month = nov,
    year = "2025",
    address = "Suzhou, China",
    publisher = "Association for Computational Linguistics",
    url = "https://aclanthology.org/2025.findings-emnlp.859/",
    doi = "10.18653/v1/2025.findings-emnlp.859",
    pages = "15880--15894",
    ISBN = "979-8-89176-335-7"
}

@inproceedings{yang2018hotpotqa,
  title     = {{HotpotQA}: A Dataset for Diverse, Explainable Multi-hop Question Answering},
  author    = {Yang, Zhilin and Qi, Peng and Zhang, Saizheng and Bengio, Yoshua and
               Cohen, William W. and Salakhutdinov, Ruslan and Manning, Christopher D.},
  booktitle = {Proceedings of the 2018 Conference on Empirical Methods in Natural Language Processing (EMNLP)},
  pages     = {2369--2380},
  year      = {2018}
}

@article{chen2025aiopslab,
  title={{AIOpsLab}: A Holistic Framework to Evaluate {AI} Agents for Enabling Autonomous Clouds},
  author={Chen, Yinfang and Shetty, Manish and Somashekar, Gagan and Ma, Minghua and Simmhan, Yogesh and Mace, Jonathan and Bansal, Chetan and Wang, Rujia and Rajmohan, Saravan},
  journal={Proceedings of Machine Learning and Systems},
  volume={7},
  year={2025}
}

@article{drosos2024your,
  title={When Your Infrastructure Is a Buggy Program: Understanding Faults in Infrastructure as Code Ecosystems},
  author={Drosos, Georgios-Petros and Sotiropoulos, Thodoris and Alexopoulos, Georgios and Mitropoulos, Dimitris and Su, Zhendong},
  journal={Proceedings of the ACM on Programming Languages},
  volume={8},
  number={OOPSLA2},
  pages={2490--2520},
  year={2024},
  publisher={ACM New York, NY, USA}
}

@article{pahune2025transitioning,
  title={Transitioning from {MLOps} to {LLMOps}: Navigating the Unique Challenges of Large Language Models},
  author={Pahune, Saurabh and Akhtar, Zahid},
  journal={Information},
  volume={16},
  number={2},
  pages={87},
  year={2025},
  publisher={MDPI}
}

@book{brikman2022terraform,
  title={{Terraform}: Up and Running: Writing Infrastructure as Code},
  author={Brikman, Yevgeniy},
  year={2022},
  publisher={O'Reilly Media},
  edition={3rd}
}

@article{poniszewska2021kubernetes,
  doi={10.3390/s21051910},
  title={{Kubernetes} Cluster for Automating Software Production Environment},
  author={Poniszewska-Mara{\'n}da, Aneta and Czechowska, Ewa},
  journal={Sensors},
  volume={21},
  number={5},
  pages={1910},
  year={2021},
  publisher={MDPI}
}

@inproceedings{jimenez2024swe,
  title={{SWE}-bench: Can Language Models Resolve Real-World {GitHub} Issues?},
  author={Jimenez, Carlos E and Yang, John and Wettig, Alexander and Yao, Shunyu and Pei, Kexin and Press, Ofir and Narasimhan, Karthik},
  booktitle={International Conference on Learning Representations},
  year={2024}
}

@article{barua2014developers,
  doi={10.1007/s10664-012-9231-y},
  title={What Are Developers Talking About? {An} Analysis of Topics and Trends in {Stack Overflow}},
  author={Barua, Anton and Thomas, Stephen W and Hassan, Ahmed E},
  journal={Empirical Software Engineering},
  volume={19},
  number={3},
  pages={619--654},
  year={2014},
  publisher={Springer}
}

@inproceedings{samanta2024efficient,
  title={Efficient Incident Summarization in {ITOps}: Leveraging Entity-Based Grouping},
  author={Samanta, Suranjana and Chatterjee, Oishik and Gupta, Hiten and Mohapatra, Prateeti and De Magalhaes, Arthur and Rahane, Ameet and Palaci-Olgun, Marc and Kattinakere, Ragu},
  booktitle={2024 IEEE International Conference on Software Services Engineering (SSE)},
  pages={97--103},
  year={2024},
  organization={IEEE}
}

@inproceedings{samanta2023insightssumm,
  title={{InsightsSumm}: Summarization of {ITOps} Incidents through In-Context Prompt Engineering},
  author={Samanta, Suranjana and Chatterjee, Oishik and Boyette, Neil and Liu, Guangya and Mohapatra, Prateeti},
  booktitle={2023 IEEE 16th International Conference on Cloud Computing (CLOUD)},
  pages={290--292},
  year={2023},
  organization={IEEE}
}

@incollection{sarma2026aiops,
  title={{AIOps}: Evolution and Transformation of {IT} Operations},
  author={Sarma, ADN},
  booktitle={Agile AI-Powered Project Management for Modern Delivery Organizations},
  pages={27--82},
  year={2026},
  publisher={IGI Global Scientific Publishing}
}

@misc{deng2023ibm,
  title={{IBM} Intelligent Remediation for {ITOps} Incidents Powered by Generative {AI}},
  author={Deng, Yu},
  howpublished={Expo Demonstration, Conference on Neural Information Processing Systems ({NeurIPS})},
  url={https://neurips.cc/virtual/2023/expo-demonstration/78270},
  year={2023}
}

@article{chandrachood2023optimizing,
  title={Optimizing Resource Allocation through Telemetry-Based Performance Monitoring},
  author={Chandrachood, Akshay},
  journal={North American Journal of Engineering Research},
  volume={4},
  number={4},
  year={2023}
}

@article{wang2024netconfeval,
  title={{NetConfEval}: Can {LLMs} Facilitate Network Configuration?},
  author={Wang, Changjie and Scazzariello, Mariano and Farshin, Alireza and Ferlin, Simone and Kosti{\'c}, Dejan and Chiesa, Marco},
  journal={Proceedings of the ACM on Networking},
  volume={2},
  number={CoNEXT2},
  pages={1--25},
  year={2024},
  publisher={ACM New York, NY, USA}
}

@misc{panda2025benefits,
  doi={10.2139/ssrn.5285768},
  title={Benefits of Site Reliability Engineering ({SRE}) in Modern Technology Environments},
  author={Panda, Sibaram Prasad and Koneti, Subramanya Bharathvamsi and Muppala, Mohanraju},
  howpublished={SSRN Working Paper 5285768},
  year={2025}
}

@inproceedings{chigurupati2025site,
  title={Ensuring Reliability in Quantum Computing and Cryptography Through Site Reliability Engineering ({SRE}) Principles},
  author={Chigurupati, Mourya and Toorpu, Arvind Reddy and Anand, Karanveer and Jagtap, Ashwini},
  booktitle={Innovative Computing and Communications: Proceedings of ICICC 2025},
  volume={2},
  series={Lecture Notes in Networks and Systems},
  pages={153--163},
  year={2025},
  publisher={Springer Nature Singapore},
  doi={10.1007/978-981-96-6681-2_12}
}

@incollection{sarma2027ai,
  title={{AI} for {IT} Operations ({AIOps}): Applications, Challenges, and Benefits},
  author={Sarma, ADN},
  booktitle={Encyclopedia of Modern Artificial Intelligence},
  pages={1644--1675},
  year={2026},
  doi={10.4018/407628},
  publisher={IGI Global Scientific Publishing}
}

@article{zhou2024haicosystem,
  title={{HAICOSYSTEM}: An Ecosystem for Sandboxing Safety Risks in Human-{AI} Interactions},
  author={Zhou, Xuhui and Kim, Hyunwoo and Brahman, Faeze and Jiang, Liwei and Zhu, Hao and Lu, Ximing and Xu, Frank and Lin, Bill Yuchen and Choi, Yejin and Mireshghallah, Niloofar and others},
  journal={arXiv preprint arXiv:2409.16427},
  year={2024}
}

@inproceedings{malik2026edge,
  title={Edge-Enabled Intelligence: Leveraging Cloudless {AI} and {AIOps} Frameworks for Advanced Fraud Detection in Retail Transactions},
  author={Rana, Milankumar and Malik, Monika and Giri, Nandita},
  booktitle={Proceedings of Fifth International Conference on Computing and Communication Networks: ICCCN 2025, Volume 10},
  volume={10},
  series={Lecture Notes in Networks and Systems},
  pages={196--206},
  year={2026},
  doi={10.1007/978-3-032-14194-1_15},
  organization={Springer Nature}
}

@incollection{rana2026aiops,
  doi={10.1002/9781394382484.ch9},
  title={{AIOps}: Intelligent Cloud Observability and Incident Management},
  author={Rana, Milankumar and Shah, Jyoti Kunal},
  booktitle={Next-Generation Technologies in Cloud Computing: From {AI} and Security to Sustainability},
  pages={151--186},
  year={2026},
  publisher={Wiley}
}

@inproceedings{xu2025openrca,
  title={{OpenRCA}: Can Large Language Models Locate the Root Cause of Software Failures?},
  author={Xu, Junjielong and Zhang, Qinan and Zhong, Zhiqing and He, Shilin and Zhang, Chaoyun and Lin, Qingwei and Pei, Dan and He, Pinjia and Zhang, Dongmei and Zhang, Qi},
  booktitle={International Conference on Learning Representations},
  year={2025}
}

@article{pham2025rcaeval,
  title={{RCAEval}: A Benchmark for Root Cause Analysis of Microservice Systems with Telemetry Data},
  author={Pham, Luan and Zhang, Hongyu and Ha, Huong and Salim, Flora and Zhang, Xiuzhen},
  journal={arXiv preprint arXiv:2412.17015},
  year={2025}
}

@article{wang2026cloudopsbench,
  title={{Cloud-OpsBench}: A Reproducible Benchmark for Agentic Root Cause Analysis in Cloud Systems},
  author={Wang, Yilun and Yu, Guangba and Huang, Haiyu and Wang, Zirui and Huang, Yujie and Chen, Pengfei and Lyu, Michael R.},
  journal={arXiv preprint arXiv:2603.00468},
  year={2026}
}

@article{cui2024logeval,
  title={{LogEval}: A Comprehensive Benchmark Suite for Large Language Models in Log Analysis},
  author={Cui, Tianyu and Ma, Shiyu and Chen, Ziang and Xiao, Tong and Tao, Shimin and Liu, Yilun and Zhang, Shenglin and Lin, Duoming and Liu, Changchang and Cai, Yuzhe and Meng, Weibin and Sun, Yongqian and Pei, Dan},
  journal={arXiv preprint arXiv:2407.01896},
  year={2024}
}

@inproceedings{clark2026sregym,
  title={{SREGym}: A Live Benchmark for {AI} {SRE} Agents with High-Fidelity Failure Scenarios},
  author={Clark, Jackson and Su, Yiming and Pial, Saad Mohammad Rafid and Tian, Yifang and Gniedziejko, Lily and Jacobsen, Hans-Arno and Chen, Yinfang and Xu, Tianyin},
  booktitle={Advances in Neural Information Processing Systems},
  year={2026}
}

@inproceedings{guo2024owl,
  title     = {{OWL}: A Large Language Model for {IT} Operations},
  author    = {Guo, Hongcheng and Yang, Jian and Liu, Jiaheng and Yang, Liqun and Chai, Linzheng and Bai, Jiaqi and Peng, Junran and Hu, Xiaorong and Chen, Chao and Zhang, Dongfeng and Shi, Xu and Zheng, Tieqiao and Zheng, Liangfan and Zhang, Bo and Xu, Ke and Li, Zhoujun},
  booktitle = {The Twelfth International Conference on Learning Representations ({ICLR})},
  year      = {2024},
  url       = {https://openreview.net/forum?id=SZOQ9RKYJu}
}

@inproceedings{ahmed2023recommending,
  title     = {Recommending Root-Cause and Mitigation Steps for Cloud Incidents using Large Language Models},
  author    = {Ahmed, Toufique and Ghosh, Supriyo and Bansal, Chetan and Zimmermann, Thomas and Zhang, Xuchao and Rajmohan, Saravan},
  booktitle = {Proceedings of the 45th IEEE/ACM International Conference on Software Engineering ({ICSE})},
  pages     = {1737--1749},
  publisher = {IEEE},
  year      = {2023},
  doi       = {10.1109/ICSE48619.2023.00149}
}

@inproceedings{balepur2024artifacts,
  title     = {Artifacts or Abduction: How Do {LLM}s Answer Multiple-Choice Questions Without the Question?},
  author    = {Balepur, Nishant and Ravichander, Abhilasha and Rudinger, Rachel},
  booktitle = {Proceedings of the 62nd Annual Meeting of the Association for Computational Linguistics (Volume 1: Long Papers)},
  pages     = {10308--10330},
  address   = {Bangkok, Thailand},
  publisher = {Association for Computational Linguistics},
  year      = {2024},
  doi       = {10.18653/v1/2024.acl-long.555},
  url       = {https://aclanthology.org/2024.acl-long.555/}
}

@inproceedings{panickssery2024llm,
  title     = {{LLM} Evaluators Recognize and Favor Their Own Generations},
  author    = {Panickssery, Arjun and Bowman, Samuel R. and Feng, Shi},
  booktitle = {Advances in Neural Information Processing Systems 37 ({NeurIPS})},
  volume    = {37},
  pages     = {68772--68802},
  publisher = {Curran Associates, Inc.},
  year      = {2024},
  doi       = {10.52202/079017-2197},
  url       = {https://proceedings.neurips.cc/paper_files/paper/2024/hash/7f1f0218e45f5414c79c0679633e47bc-Abstract-Conference.html}
}

@inproceedings{zheng2023judging,
  title     = {Judging {LLM}-as-a-Judge with {MT-Bench} and {Chatbot Arena}},
  author    = {Zheng, Lianmin and Chiang, Wei-Lin and Sheng, Ying and Zhuang, Siyuan and Wu, Zhanghao and Zhuang, Yonghao and Lin, Zi and Li, Zhuohan and Li, Dacheng and Xing, Eric P. and Zhang, Hao and Gonzalez, Joseph E. and Stoica, Ion},
  booktitle = {Advances in Neural Information Processing Systems 36 ({NeurIPS}), Datasets and Benchmarks Track},
  volume    = {36},
  pages     = {46595--46623},
  publisher = {Curran Associates, Inc.},
  year      = {2023},
  doi       = {10.52202/075280-2020},
  url       = {https://proceedings.neurips.cc/paper_files/paper/2023/hash/91f18a1287b398d378ef22505bf41832-Abstract-Datasets_and_Benchmarks.html}
}

@inproceedings{gururangan2018annotation,
  title     = {Annotation Artifacts in Natural Language Inference Data},
  author    = {Gururangan, Suchin and Swayamdipta, Swabha and Levy, Omer and Schwartz, Roy and Bowman, Samuel R. and Smith, Noah A.},
  booktitle = {Proceedings of the 2018 Conference of the North American Chapter of the Association for Computational Linguistics: Human Language Technologies, Volume 2 (Short Papers)},
  pages     = {107--112},
  address   = {New Orleans, Louisiana},
  publisher = {Association for Computational Linguistics},
  year      = {2018},
  url       = {https://aclanthology.org/N18-2017/}
}

\newpage
\appendix
\renewcommand{\topfraction}{0.9}\renewcommand{\dbltopfraction}{0.9}
\renewcommand{\textfraction}{0.07}\renewcommand{\floatpagefraction}{0.7}\renewcommand{\dblfloatpagefraction}{0.7}
\setcounter{topnumber}{4}\setcounter{dbltopnumber}{4}\setcounter{totalnumber}{6}
\section*{Appendix Overview}
The appendices are grouped in four parts. Each part lists its sections below with a short description.

\paragraph{Background.}
Appendix~\ref{app:related} covers AIOps and LLM-for-ITOps work beyond the benchmarks in \S\ref{sec:related}. Appendix~\ref{app:discovery} covers the search procedure and inclusion criteria for the benchmark comparison. Appendix~\ref{app:requirements} covers the four benchmark requirements and ORBIT's coverage of each.

\paragraph{Benchmark construction and data.}
Appendix~\ref{app:construction} covers sources, indexing, retrieval, generation, validation, and cleaning, with the taxonomy and pipeline figures. Appendix~\ref{app:ts} covers selection, generation, audit, and rubrics for the rebuilt troubleshooting items. Appendix~\ref{app:labels} covers how hard labels and cohorts were assigned. Appendix~\ref{app:examples} covers annotated items across formats and tiers, with their known weaknesses.

\paragraph{Evaluation details and additional results.}
Appendix~\ref{app:protocol} covers models, prompts, judge prompt, and decoding settings. Appendix~\ref{app:stats} covers every significance test, accuracy by reasoning tier, and the detectable-difference formula. Appendix~\ref{app:allitems} covers accuracy on all generated items under self-judging and the independent judge. Appendix~\ref{app:oldpass} covers the earlier self-judged passes, kept for transparency.

\paragraph{Validity analyses.}
Appendix~\ref{app:iaa} covers agreement among experts, among LLM graders, and between the two groups. Appendix~\ref{app:external} covers the topic overlap with Stack Overflow. Appendix~\ref{app:shortcut} covers surface-cue probes over options and questions. Appendix~\ref{app:contamination} covers planned protocols for measuring memorization.

\section{Extended Related Work}
\label{app:related}

This appendix extends \S\ref{sec:related} by providing more detail on earlier work on AIOps and LLMs for IT operations.

\paragraph{From ITOps to AIOps.}
ITOps keeps enterprise infrastructure available, fast, and reliable through monitoring, incident management, configuration, and service delivery. Cloud-native scale has outgrown manual work, and AIOps adds machine learning for anomaly detection, root-cause analysis, predictive maintenance, and automated fixes. \citet{dang2019aiops} summarize Microsoft's AIOps experience and cite poor data quality, missing labels, and complex dependencies. \citet{zheng2023survey} survey 134 studies and find that code-specialized LLMs outperform general-purpose LLMs on code tasks.

\paragraph{LLMs for ITOps.}
LLMs help with configuration generation, API evaluation, and root-cause analysis \citep{wang2024netconfeval}, yet offline scores and real reliability diverge. For Kubernetes manifest generation, LoRA trains 3$\times$ faster with 45\% less memory, while full fine-tuning keeps higher accuracy \citep{sacco2025intent}. GraphQLRestBench tests sequential API calls, but its non-executable setting limits transfer \citep{saha2024sequential}. RCACopilot reaches 0.95 BERTScore F1 with few-shot retrieval against 0.66 zero-shot, so grounding helps, but its gain hinges on good retrieval quality \citep{shan2025rca}.

\section{Benchmark Discovery}
\label{app:discovery}

This appendix records how we found the benchmarks in Table~\ref{tab:benchmark_comparison}, what we included, and why we compare task properties rather than scores.

\paragraph{Search procedure.}
We ran a scoping search on 22 October 2026 over a general web search engine, arXiv, and the ICLR, ACM, and NeurIPS proceedings. Queries combined a task term with an evaluation term, such as ``LLM benchmark root cause analysis'', ``benchmark AI agents site reliability engineering'', ``LLM benchmark log analysis'', and ``LLM benchmark incident management cloud operations''. We also followed the related work of the benchmarks found. We did not search the ACL Anthology or the ICSE, FSE, and ASE proceedings separately, so some benchmarks may be missing.

\paragraph{Inclusion criteria.}
We included a resource if it (i) releases an evaluation set or environment, (ii) targets IT operations, SRE, incident management, log analysis, or root-cause localization, and (iii) evaluates LLMs or LLM agents or provides baselines for root-cause methods. We excluded systems papers without a released benchmark, surveys, methods such as log parsers, and commercial products whose evaluation data are not public. Table~\ref{tab:benchmark_comparison} lists all included benchmarks along with their settings.

\paragraph{Industrial systems.}
Commercial observability platforms now ship LLM assistants for incident triage and root-cause hints. Their models and evaluation data are private, so their accuracy cannot be compared or audited, and the incidents they learn from cannot be released. ORBIT does not compete with these systems. It offers an open, source-linked test that the models behind such assistants can be run on before they take on-call duties.

\paragraph{Scope of the search.}
This is a scoping search, not a full systematic review. Gray literature is covered only where vendors publish benchmarks. We record the search strings and inclusion criteria so the search can be repeated and extended.

\paragraph{Why no direct score comparison.}
The included benchmarks score different outputs, namely executed actions, localized root causes, parsed logs, or answers. A model's ITBench success rate and its ORBIT accuracy are therefore not on a common scale. Table~\ref{tab:benchmark_comparison} instead compares task properties. Running one set of models across ORBIT, OpsEval, and OpenRCA is left to future work.

\section{Benchmark Requirements}
\label{app:requirements}

This appendix defines the four requirements behind Table~\ref{tab:benchmark_comparison} and states how far ORBIT meets each.

\paragraph{Scope.}
Operational diagnosis is the debugging of deployed software, its configuration, its infrastructure, and related processes \citep{drosos2024your}. ORBIT covers SRE tasks such as fault localization, diagnosis, and choosing remediation steps in the right order \citep{chen2025aiopslab, jhaitbench}.

\paragraph{Operational grounding.}
Each question should come from material that SREs actually produce and use, such as vendor and APM documentation, operational wikis, and incident discussions. Items made up without such material may be unrealistic or trivial. We treat grounding as a recorded property, so each grounded item keeps a reference to its source, which a reviewer with access can check.

\paragraph{Multi-entity reasoning.}
A question should require combining several pieces of evidence, not retrieving one stated fact. Real diagnosis chains observations across interacting components (\S\ref{sec:intro}), and a benchmark dominated by single-fact recall overstates operational skill. ORBIT meets this requirement for its reasoning-heavy tier and diagnostic cohort, not for every item in the corpus.

\paragraph{Evidence traceability.}
Each item should link its answer to the supporting source passages and to the reasoning used to reach it. This lets others check that the answer rests on documented evidence and find which reasoning step fails, rather than only scoring the final answer as right or wrong.

\paragraph{Technology diversity.}
The benchmark should cover the technologies of real ITOps, such as cloud platforms, infrastructure-as-code, observability tools, container orchestration, databases, and networking. Broad coverage avoids overrating a model that knows one stack. ORBIT does not yet measure coverage per technology, so this requirement is a goal rather than a verified property.

\FloatBarrier
\section{Construction Pipeline Details}
\label{app:construction}

This appendix details each stage of the pipeline in \S\ref{sec:construction}, the sources behind grounded items, and the cleaning of the raw export into the release.

\begin{figure*}[t]
  \centering
  \includegraphics[width=0.98\textwidth]{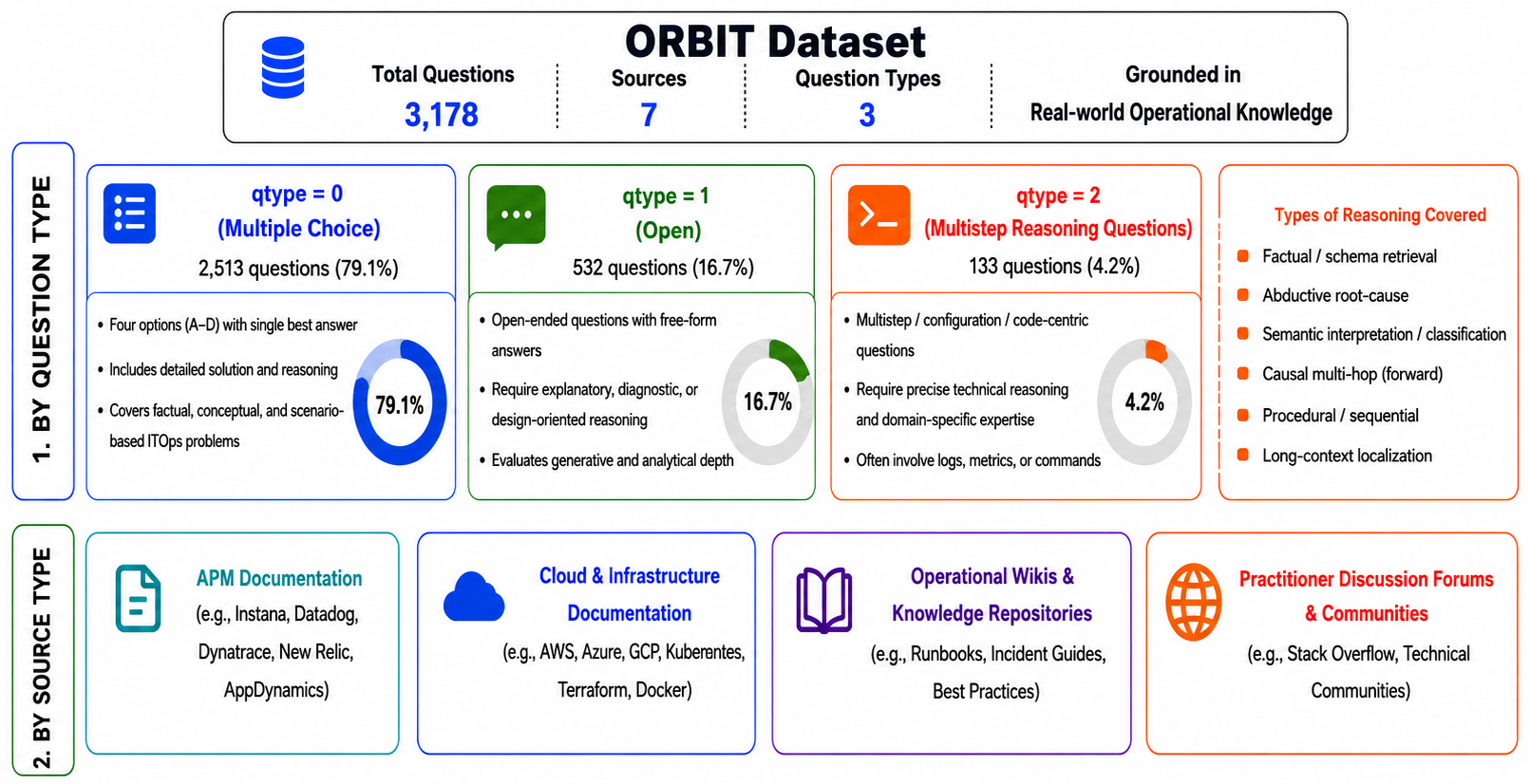}
  \caption{Taxonomy of ORBIT by question type, source type, and reasoning type.}
  \label{fig:taxonomy}
\end{figure*}

Figure~\ref{fig:taxonomy} groups seven source collections into four source types. In the figure, qtype 0, 1, and 2 are multiple-choice, open-ended, and troubleshooting items, and ``Multistep Reasoning Questions'' are the troubleshooting items. Reddit is a collected source, but no released item is grounded in it, and long-context localization has only six items.

\paragraph{Knowledge source collection.}
Operational knowledge is spread across several kinds of material. ORBIT draws on four groups. (i) APM documentation covers diagnostic procedures, telemetry meaning, alert definitions, and troubleshooting steps for Instana, Dynatrace, and Datadog. (ii) Cloud and infrastructure documentation forms the second group. (iii) Operational wikis and runbooks include incident-response guides, best practices, and an internal documentation repository. (iv) Practitioner forums such as Stack Overflow capture diagnostic habits, failure patterns, and the fixes that worked in practice.

\paragraph{Ingestion, normalization, and indexing.}
The sources differ in structure, so one ingestion pipeline fetches them, removes page formatting, and keeps code, configuration files, procedures, tables, and document hierarchies. The cleaned corpus is split into sentence-aware passages of about 500 characters with small overlap. Passages are encoded with \texttt{gemini-embedding-001} (3,072 dimensions) and indexed in a Postgres vector store. Each passage keeps its source, so items keep their context.

\paragraph{Retrieval-grounded evidence construction.}
To ground items in documents rather than model memory, ORBIT uses retrieval-grounded generation. For a target topic, semantic queries retrieve passages from documentation, which holds system facts, and from practitioner discussions, which hold diagnostic habits and troubleshooting patterns. A query returns the 8 most similar passages by cosine similarity, and a Stack Overflow search returns up to 5 questions with their 2 top answers each. The annotator picks from these, and the chosen passages form the evidence package, whose identifiers are stored with the item for later checking.

\paragraph{Task specification and generation.}
ORBIT supports multiple-choice and open-ended questions, troubleshooting scenarios, and multi-hop problems. Each task type fixes the target reasoning, output structure, and complexity. The drafting model receives the evidence package with the task rules and writes items whose answers must follow from that evidence. The tool offers \texttt{gemini-3-flash-preview} (default) and \texttt{gemini-2.5-flash}, and locally served open models. The export does not record the drafting model. Each item has a problem statement, answer options when needed, a reference solution, a reasoning trace, and coverage and source metadata.

\paragraph{Diversity and provenance controls.}
To limit topic imbalance, repetition, and weak traceability, a concept registry tracks concepts, technologies, failure modes, and reasoning patterns that have already been covered. Coverage tracking steers generation to rare areas, and topic rules limit near-duplicate scenarios. Every item keeps its evidence throughout.

\paragraph{Validation and consensus.}
Three of the authors reviewed each candidate against four criteria, namely correctness, reasoning coherence, operational validity, and answer-evidence fit. Each candidate was then accepted, revised, or rejected. Disagreements flagged disputed items, majority vote settled routine conflicts, and expert adjudication settled harder ones. Item identifiers, evidence links, review states, and revision histories were kept throughout. The released export keeps an explicit annotator verdict for 298 items, all accepted, and omits the review state of the other items.

\paragraph{Sources and technologies.}
The 2,005 internally grounded items cite 1,678 schema pages from an internal wiki, 266 saved copies of public vendor API documentation, 57 other wiki, glossary, and product pages, and 4 document titles without a path. By technology, Instana grounds about 1,360 items, seven other enterprise IT products about 570, and AWS, Azure, and GCP about 25. The 798 public items cite 794 Stack Overflow threads and 4 Wikipedia pages. Of the 375 ungrounded items, 175 are template variants with a fixed phrase.

Table~\ref{tab:orbit_feature_statistics} gives structural statistics for the 3,178 deduplicated items, where lengths count whitespace tokens, and only the mean is recorded for evidence context length. Averages are dominated by multiple-choice items, which make up 79\% of the corpus and have the shortest questions.

\paragraph{Corpus cleaning.}
The raw export held 3,350 rows. Removing 172 exact duplicates of 135 identifiers left 3,178 unique items. Rule-based checks rebuilt 107 multiple-choice option arrays that mismatched their lettered fields and set aside 57 items with duplicate correct options. Llama 3.3 70B then rewrote 470 templated open-ended answers from their solutions and made new distractors for 113 multiple-choice items, including the 57, keeping the validated answer at its original letter. Every change is logged per item. These rewrites came after expert validation and were not re-reviewed.

\begin{table}[!ht]
\centering
\small
\setlength{\tabcolsep}{4pt}
\resizebox{\columnwidth}{!}{%
\begin{tabular}{lcc}
\toprule
\textbf{Metric} & \textbf{Value} & \textbf{Unit} \\
\midrule
Question length & $14.28 \pm 5.56$ & whitespace tokens \\
Technical terms per question & $3.19 \pm 1.72$ & count \\
Answer length (option text for MCQ) & $13.72 \pm 15.66$ & whitespace tokens \\
Reference-solution length & $27.56 \pm 16.82$ & words \\
Evidence context length & $135.22$ & tokens \\
Grounding references per item & $0.98 \pm 0.46$ & count \\
Items with $\geq$2 references & $9.4$ & \% \\
\bottomrule
\end{tabular}}
\caption{Structural statistics of ORBIT, as mean $\pm$ standard deviation.}
\label{tab:orbit_feature_statistics}
\end{table}

\section{Troubleshooting Items from Practitioner Threads}
\label{app:ts}

This appendix describes how the 208 troubleshooting items in the evaluation set were selected from practitioner threads, generated, audited, and scored.

\paragraph{Thread selection.}
The forum collection holds 2,750 threads with an accepted answer (1,443 from Stack Overflow, 1,053 from Server Fault, and 254 from DevOps Stack Exchange). We kept threads whose LLM-assigned primary reasoning type is abductive root cause, that no existing ORBIT item cites, whose accepted answer has 300 to 8,000 characters and a score of at least 3, and that contain no images. We sampled 438 of them in proportion to each site, with at most 12 per leading tag. The threads were posted between 2009 and 2022.

\paragraph{Generation and audit.}
Claude Opus 5.5 received the thread title, question, and accepted answer at temperature 0. It wrote a scenario of 60 to 220 words from an operator's point of view that keeps error messages, log lines, versions, configuration fragments, and earlier attempts but does not state or hint at the cause. It also wrote a root cause and a fix taken only from the accepted answer, and two to four key points that a correct diagnosis must contain. It could mark a thread as unusable when it was not a single-cause diagnosis problem, when the answer provided only a workaround, or when the evidence could not be kept without revealing the cause. GPT-5.5 then received the same thread and the item and judged four properties independently, namely faithful, no leak, solvable, and single cause. An item was kept only if all four were held.

The generator marked 90 threads unusable. Of the 348 audited items, 140 failed, 111 of them because the scenario revealed the cause, 17 for more than one cause, 12 for an unfaithful reference, and 11 as not solvable from the scenario (an item can fail several checks). The 208 kept items come from Stack Overflow (136), Server Fault (67), and DevOps Stack Exchange (5). Docker, PostgreSQL, Kubernetes, Redis, Elasticsearch, and Nginx are the most common technologies. Scenarios average 152 words, against 25.6 words for the templated troubleshooting items they replace in ORBIT.

\paragraph{Prompt and scoring.}
Each question ends with ``What is the most likely root cause, and how would you fix it?'' The reference answer provides the root cause and the fix, and the key points serve as the judge's reference. Scoring uses the judge and a strict prompt for open-ended items.

Inspecting verdicts showed that the strict prompt fails on answers that name the correct cause and a working fix but omit a secondary detail of the reference, such as a second package to install. We therefore also graded every answer with a root-cause rubric that asks only for the same underlying cause and a fix that would resolve it (Table~\ref{tab:ts_rubrics}). We wrote this rubric after seeing the strict verdicts. It accepted 116 answers; the strict prompt rejected none. A spot check found most new acceptances correct, but some kept a main fix that would not work. The two scores bound troubleshooting accuracy from below and above.

\begin{table}[!ht]
\centering
\small
\setlength{\tabcolsep}{4pt}
\begin{tabular}{lcc}
\toprule
\textbf{Model} & \textbf{Strict} & \textbf{Root cause} \\
\midrule
Llama 3.3 70B & 16.8 (12.4--22.5) & 28.8 (23.1--35.3) \\
Mistral Small 3.1 & 13.5 (9.5--18.8) & 21.2 (16.2--27.2) \\
Llama 4 Maverick & 26.9 (21.4--33.3) & 40.9 (34.4--47.7) \\
Granite 4.0 H & 12.5 (8.7--17.7) & 18.3 (13.6--24.1) \\
Mistral Medium & 16.8 (12.4--22.5) & 33.2 (27.1--39.8) \\
\bottomrule
\end{tabular}
\caption{Troubleshooting accuracy (\%) on the rebuilt items under the strict prompt and under the root-cause rubric, with Wilson confidence intervals.}
\label{tab:ts_rubrics}
\end{table}

\paragraph{Example.}
TSF-0137 (from Server Fault). A CentOS 7 server accepts SSH logins, but SFTP sessions close right after authentication with exit status 127. The scenario lists the \texttt{sshd\_config} file, which contains the line \texttt{Subsystem sftp /usr/lib/openssh/sftp-server}. The cause is that this Debian-style path does not exist on CentOS, so the subsystem command is missing. The fix is to point the directive at the installed binary or use \texttt{internal-sftp} and then restart \texttt{sshd}.

\paragraph{Open checks.}
Experts have not yet reviewed these items. A 30-item sample is prepared, and failing items will be removed before release. The public threads are older than every evaluated model, so models may have seen them in training (\S\ref{sec:limitations}).

\section{Difficulty Labels and Cohorts}
\label{app:labels}

This appendix explains how difficulty labels and failure-mode cohorts were assigned and how they map onto the evaluation set used in \S\ref{sec:experiments}.

\paragraph{Hard labels.}
Hard items are not perturbed copies of standard items. Each item receives a difficulty score that combines a rule-based score with an LLM review of reasoning, evidence, and troubleshooting complexity, and any item above a fixed threshold is labeled hard. This gives 740 hard items (478 open-ended, 133 troubleshooting, 129 multiple-choice) and 2,438 standard items. Because selection is by difficulty, the gap shows how sharply accuracy falls on the hardest items, not robustness to a controlled perturbation.

\paragraph{Failure-mode cohorts.}
During construction, an LLM gives each item a primary reasoning type and a failure-mode cohort. In the evaluation set, the diagnostic cohort contains 362 multiple-choice items, and the multi-hop causal cohort contains 235 open-ended and 16 multiple-choice items. The excluded template items held all other diagnostic items. Cohort labels, like reasoning-type labels, are not checked against human labels.

\section{Representative Instances}
\label{app:examples}

Table~\ref{tab:orbit_dataset_preview} shows four items across tiers, with their LLM-assigned primary reasoning type. Questions, options, and summaries are shortened. The PHP-FPM item's full question states that the investigation used Instana traces and packet captures. DP-00041 is in the retrieval tier, and DP-00029 and DP-00025 are in the semantic-interpretation tier.

Two of the four items show known weak spots. DP-00029 and DP-00025 ask what an entry such as \texttt{.instana} or \texttt{.gcp} is in a documentation hierarchy, which tests one repository's layout rather than operations. In all three, only the correct option uses code formatting, a cue present in 102 multiple-choice items. We show them so that users see the whole range of item quality in the corpus.

\begin{table*}[t]
\centering
\scriptsize
\setlength{\tabcolsep}{4pt}
\renewcommand{\arraystretch}{1.2}
\begin{tabularx}{\textwidth}{>{\raggedright\arraybackslash}p{1.7cm}*{4}{>{\raggedright\arraybackslash}X}}
\toprule
& \textbf{inst-101} & \textbf{DP-00029} & \textbf{DP-00041} & \textbf{DP-00025} \\
\midrule
\textbf{Category} & Troubleshooting (MCQ) & Product Understanding & Schema Extraction & Documentation Navigation \\
\textbf{Domain} & Observability & APM (Instana) & AWS (Cloud Provider) & GCP (Cloud Provider) \\
\textbf{Reasoning type} & Abductive root cause & Semantic interpretation & Factual / schema retrieval & Semantic interpretation \\
\textbf{Grounding source} & Community Discussion & Product Documentation & API Reference & Product Documentation \\
\midrule
\textbf{Question} &,
PHP-FPM requests intermittently hang for 20--60 seconds after output generation. What monitoring-related root cause was identified? &
Within the Instana documentation hierarchy, what functional purpose does \texttt{.instana} serve? &
What is the complete field structure of \texttt{AwsEc2Instance}? &
Based on the documentation hierarchy, what type of entry is \texttt{.gcp}? \\
\addlinespace
\textbf{Options} &
(A) UDP metric corruption, (B) TCP metric transmission contention, (C) OOM-killed collector, (D) Missing OTel correlation &
(A) IBM Instana observability platform, (B) Deprecated parameter, (C) Monitoring latency field, (D) Authentication schema &
(A) 3 fields (\texttt{instance\_id}, \ldots), (B) 7 fields (\texttt{aws\_resource}, \ldots), (C) 4 fields (\texttt{image\_id}, \ldots), (D) 5 fields (\texttt{aws\_resource}, \texttt{monitoring}, \texttt{private\_ip}, \texttt{tenancy}, \texttt{virtual\_machine}) &
(A) Category page, (B) Deprecated parameter, (C) Monitoring component, (D) Google Cloud Platform infrastructure resources \\
\addlinespace
\textbf{Answer} & B & A & D & D \\
\addlinespace
\textbf{Reasoning summary} &
Correlates application traces, packet captures, and monitoring behavior. Resource contention during Instana TCP metric transmission delayed request completion, although PHP execution had finished. &
Requires understanding of product taxonomy and documentation structure. The entry represents the Instana platform entity rather than a schema attribute or configuration parameter. &
Requires retrieval of the exact schema definition from AWS documentation and differentiation from related infrastructure resource types. &
Requires semantic interpretation of documentation organization and recognition of GCP as a cloud infrastructure product category. \\
\bottomrule
\end{tabularx}
\caption{Representative ORBIT items across domains and reasoning tiers, with shortened questions, options, and reasoning summaries.}
\label{tab:orbit_dataset_preview}
\end{table*}

\begin{table*}[t]
\centering
\small
\renewcommand{\arraystretch}{1.15}
\begin{tabularx}{\textwidth}{p{1.6cm}X}
\toprule
\textbf{Field} & \textbf{DP-02007 (troubleshooting format, multi-hop cohort, abductive root cause, hard)} \\
\midrule
Question & Given the following production scenario, perform root cause analysis. Scenario: A Kubernetes pod running a Java application is continuously OOMKilled. The node has 8GB of RAM. The pod has memory requests and limits set to 256Mi. The Java application is started with \texttt{-Xmx512m -Xms128m}. \\
Reference solution & The pod's hard memory limit (256Mi) is lower than the JVM's maximum heap size (\texttt{-Xmx512m}). Either increase the Kubernetes memory limit above 512Mi or reduce the JVM heap to fit within 256Mi. \\
Reasoning trace (excerpt) & (1) The node has ample RAM, but Kubernetes enforces memory at the container level through cgroups. (2) \texttt{limits.memory} of 256Mi is a hard ceiling, and exceeding it triggers the OOM killer. (3) The JVM is allowed to grow its heap to 512MB. (4) The configured heap exceeds the container limit, so the container is killed. \\
\bottomrule
\end{tabularx}
\caption{A full troubleshooting item as shown to models, with its reference solution and reasoning trace.}
\label{tab:diagnostic_example}
\end{table*}

Table~\ref{tab:diagnostic_example} shows a full troubleshooting item as models receive it. The question holds every fact needed, and the solution follows by combining them. It shows the format well, but it is not a clean item. Its recorded source is a document title with no resolvable path, and its answer field stores a letter from a generic option set. The judge also receives the reference solution shown here. It is textbook Kubernetes knowledge needing no recorded evidence.

\section{Evaluation Protocol Details}
\label{app:protocol}

This appendix provides the model serving, prompting, judging, and decoding settings used in \S\ref{sec:experiments}.

\paragraph{Scoring and prompting rationale.}
Exact match suits multiple-choice items because each has one keyed option. A correct generative answer can take many forms and needs a semantic check against the reference. All reported runs are zero-shot, which measures out-of-the-box behavior. Cohorts are not subsampled, so each cohort column holds every evaluation item with that label (Appendix~\ref{app:labels}).

\paragraph{Models and serving.}
All five models were served through watsonx.ai using its catalog versions, for example, \texttt{meta-llama/llama-3-3-70b-instruct}. Answers come from the chat interface, except Granite 4.0 H Small's multiple-choice answers, which come from the text interface (\S\ref{sec:experiments}). A provider-agnostic client handles prompt construction, decoding, response caching, and output collection. The harness counts an errored request or an unparseable verdict as an error and retries it. No requests failed in the reported runs. Proprietary frontier models were not evaluated.

\paragraph{Prompts.}
Multiple-choice prompts give the question and four lettered options, instruct the model to act as an IT operations engineer, and ask for a single letter, extracted with three fixed patterns. A response with no extractable letter is scored wrong. Generation is capped at 8,192 new tokens.

Open-ended and troubleshooting prompts ask for a correct, concise answer. The judge receives the question, reference answer, reference explanation, and candidate answer. Its system prompt reads ``You are a STRICT grader for IT-operations question answering. [\ldots] Mark the candidate CORRECT only if it states the same core conclusion as the reference. The specific root cause, fix, command, or value must actually match. Mark it INCORRECT if it is vague or hedged, only partially correct, omits the key point, contradicts the reference, leads with a different primary claim, or merely restates the question. Being plausible or topically related is NOT sufficient. When in doubt, answer NO.'' The verdict is the first word of the reply if it is YES or NO, and otherwise the last YES or NO. The judge may write up to 4,096 tokens.

\paragraph{Prompting settings.}
Zero-shot prompts give no examples. The troubleshooting format and the diagnostic cohort no longer overlap, since the rebuilt troubleshooting items carry no cohort label.

\paragraph{Deterministic configuration.}
Answers use greedy decoding, and the judge runs at temperature 0 with the prompt quoted above. Response caching makes reruns of the released harness return stored outputs, although provider-side updates can still change fresh outputs. We will release the cached raw responses and judge verdicts for every cell, so others can rescore with a different judge.

\section{Statistical Analysis}
\label{app:stats}

Tables~\ref{tab:stats_main} and~\ref{tab:stats_more} list every test behind \S\ref{sec:experiments}, computed from per-item outputs. The two-proportion $z$-test is a large-sample approximation that is adequate at these sample sizes. Paired tests on shared items would be more powerful. Table~\ref{tab:by_type} gives multiple-choice accuracy by reasoning tier for each model.

\begin{table}[!ht]
\centering
\small
\setlength{\tabcolsep}{3pt}
\resizebox{\columnwidth}{!}{%
\begin{tabular}{lrrr}
\toprule
\textbf{Comparison} & \textbf{$\Delta$ (pts)} & \textbf{$p_{\text{Holm}}$} & \textbf{$h$} \\
\midrule
\multicolumn{4}{l}{\textit{Format}} \\
Llama 3.3 70B, MC $\to$ OE & 52.6 & $<10^{-111}$ & 1.16 \\
Llama 3.3 70B, MC $\to$ TS & 71.3 & $<10^{-148}$ & 1.59 \\
Llama 3.3 70B, OE $\to$ TS & 18.7 & $<10^{-5}$ & 0.43 \\
Mistral Small 3.1, MC $\to$ OE & 40.7 & $<10^{-72}$ & 0.92 \\
Mistral Small 3.1, MC $\to$ TS & 74.9 & $<10^{-163}$ & 1.69 \\
Mistral Small 3.1, OE $\to$ TS & 34.2 & $<10^{-14}$ & 0.77 \\
Llama 4 Maverick, MC $\to$ OE & 40.5 & $<10^{-63}$ & 0.89 \\
Llama 4 Maverick, MC $\to$ TS & 58.9 & $<10^{-95}$ & 1.28 \\
Llama 4 Maverick, OE $\to$ TS & 18.4 & $<10^{-4}$ & 0.39 \\
Granite 4.0 H, MC $\to$ OE & 26.8 & $<10^{-24}$ & 0.59 \\
Granite 4.0 H, MC $\to$ TS & 69.0 & $<10^{-109}$ & 1.53 \\
Granite 4.0 H, OE $\to$ TS & 42.2 & $<10^{-20}$ & 0.94 \\
Mistral Medium, MC $\to$ OE & 38.3 & $<10^{-66}$ & 0.88 \\
Mistral Medium, MC $\to$ TS & 72.2 & $<10^{-158}$ & 1.62 \\
Mistral Medium, OE $\to$ TS & 33.8 & $<10^{-13}$ & 0.74 \\
\midrule
\multicolumn{4}{l}{\textit{Standard vs.\ hard multiple-choice}} \\
Llama 3.3 70B & 30.0 & $<10^{-23}$ & 0.72 \\
Mistral Small 3.1 & 33.5 & $<10^{-29}$ & 0.80 \\
Llama 4 Maverick & 29.2 & $<10^{-19}$ & 0.68 \\
Granite 4.0 H & 31.9 & $<10^{-19}$ & 0.70 \\
Mistral Medium & 30.1 & $<10^{-25}$ & 0.74 \\
\midrule
\multicolumn{4}{l}{\textit{Other vs.\ diagnostic multiple-choice}} \\
Llama 3.3 70B & 33.9 & $<10^{-74}$ & 0.85 \\
Mistral Small 3.1 & 32.3 & $<10^{-68}$ & 0.82 \\
Llama 4 Maverick & 31.2 & $<10^{-54}$ & 0.76 \\
Granite 4.0 H & 31.9 & $<10^{-45}$ & 0.72 \\
Mistral Medium & 30.1 & $<10^{-62}$ & 0.78 \\
\midrule
\multicolumn{4}{l}{\textit{OE multi-hop vs.\ troubleshooting}} \\
Llama 3.3 70B & 24.4 & $<10^{-7}$ & 0.55 \\
Mistral Small 3.1 & 41.0 & $<10^{-18}$ & 0.91 \\
Llama 4 Maverick & 26.3 & $<10^{-7}$ & 0.54 \\
Granite 4.0 H & 50.5 & $<10^{-26}$ & 1.11 \\
Mistral Medium & 41.0 & $<10^{-17}$ & 0.88 \\
\bottomrule
\end{tabular}}
\par\smallskip{\scriptsize$\Delta$ is the first accuracy minus the second, and $h$ is Cohen's $h$. Holm's correction is applied within each comparison group. MC is multiple-choice, OE is open-ended, and TS is troubleshooting.\par}
\caption{Significance tests for the comparisons behind the main findings.}
\label{tab:stats_main}
\end{table}

\begin{table}[!ht]
\centering
\small
\setlength{\tabcolsep}{3pt}
\resizebox{\columnwidth}{!}{%
\begin{tabular}{lrrr}
\toprule
\textbf{Comparison} & \textbf{$\Delta$ (pts)} & \textbf{$p_{\text{Holm}}$} & \textbf{$h$} \\
\midrule
\multicolumn{4}{l}{\textit{Model pairs, overall accuracy}} \\
L3.3 vs M-Small & $-$1.2 & 0.816 & $-$0.03 \\
L3.3 vs L4-Mav & 0.3 & 0.816 & 0.01 \\
L3.3 vs Granite & 4.0 & 0.002 & 0.09 \\
L3.3 vs M-Med & $-$2.2 & 0.169 & $-$0.05 \\
M-Small vs L4-Mav & 1.4 & 0.712 & 0.03 \\
M-Small vs Granite & 5.1 & $<10^{-4}$ & 0.12 \\
M-Small vs M-Med & $-$1.1 & 0.816 & $-$0.03 \\
L4-Mav vs Granite & 3.7 & 0.005 & 0.09 \\
L4-Mav vs M-Med & $-$2.5 & 0.107 & $-$0.06 \\
Granite vs M-Med & $-$6.2 & $<10^{-7}$ & $-$0.15 \\
\midrule
\multicolumn{4}{l}{\textit{Model pairs, open-ended}} \\
L3.3 vs M-Small & $-$12.1 & 0.022 & $-$0.25 \\
L3.3 vs L4-Mav & $-$9.7 & 0.109 & $-$0.20 \\
L3.3 vs Granite & $-$19.1 & $<10^{-4}$ & $-$0.39 \\
L3.3 vs M-Med & $-$15.1 & 0.002 & $-$0.31 \\
M-Small vs L4-Mav & 2.3 & 0.975 & 0.05 \\
M-Small vs Granite & $-$7.0 & 0.426 & $-$0.14 \\
M-Small vs M-Med & $-$3.0 & 0.975 & $-$0.06 \\
L4-Mav vs Granite & $-$9.4 & 0.131 & $-$0.19 \\
L4-Mav vs M-Med & $-$5.4 & 0.758 & $-$0.11 \\
Granite vs M-Med & 4.0 & 0.975 & 0.08 \\
\midrule
\multicolumn{4}{l}{\textit{Model pairs, troubleshooting}} \\
L3.3 vs M-Small & 3.4 & 1.000 & 0.09 \\
L3.3 vs L4-Mav & $-$10.1 & 0.102 & $-$0.25 \\
L3.3 vs Granite & 4.3 & 1.000 & 0.12 \\
L3.3 vs M-Med & 0.0 & 1.000 & 0.00 \\
M-Small vs L4-Mav & $-$13.5 & 0.006 & $-$0.34 \\
M-Small vs Granite & 1.0 & 1.000 & 0.03 \\
M-Small vs M-Med & $-$3.4 & 1.000 & $-$0.09 \\
L4-Mav vs Granite & 14.4 & 0.002 & 0.37 \\
L4-Mav vs M-Med & 10.1 & 0.102 & 0.25 \\
Granite vs M-Med & $-$4.3 & 1.000 & $-$0.12 \\
\bottomrule
\end{tabular}}
\par\smallskip{\scriptsize Holm's correction is applied within each block. L3.3 is Llama 3.3 70B, L4-Mav is Llama 4 Maverick, M-Small is Mistral Small 3.1, and M-Med is Mistral Medium.\par}
\caption{Significance tests between each pair of models on the evaluation set.}
\label{tab:stats_more}
\end{table}

\begin{table*}[t]
\centering
\small
\setlength{\tabcolsep}{3pt}
\begin{tabular}{lccc}
\toprule
\textbf{Model} & \textbf{Retrieval} & \textbf{Semantic} & \textbf{Reasoning-heavy} \\
& ($n{=}1{,}986$) & ($n{=}144$) & ($n{=}383$) \\
\midrule
Llama 3.3 70B & 97.0 (96.2--97.7) & 47.2 (39.2--55.3) & 57.4 (52.4--62.3) \\
Mistral Small 3.1 & 97.1 (96.2--97.7) & 49.3 (41.3--57.4) & 58.0 (53.0--62.8) \\
Llama 4 Maverick & 94.2 (93.1--95.2) & 43.1 (35.3--51.2) & 58.5 (53.5--63.3) \\
Granite 4.0 H & 90.4 (89.1--91.6) & 31.2 (24.2--39.2) & 53.8 (48.8--58.7) \\
Mistral Medium & 96.8 (95.9--97.5) & 50.7 (42.6--58.7) & 63.2 (58.2--67.9) \\
\bottomrule
\end{tabular}
\caption{Zero-shot multiple-choice accuracy (\%) by tier of the LLM-assigned primary reasoning type, with Wilson confidence intervals.}
\label{tab:by_type}
\end{table*}

\paragraph{Minimum detectable difference.}
For two models with accuracy $p$ on $n$ items each, an unpaired two-sided test at $\alpha=0.05$ with 80\% power detects a difference of about $2.8\sqrt{2p(1-p)/n}$. With $p=0.78$ and $n=3{,}019$ this is 3.0 points. With $p=0.47$ and $n=298$ (open-ended), it is 11.4 points, and with $p=0.17$ and $n=208$ (troubleshooting), it is 10.4 points. Holm's correction raises these detectable differences further.

\section{Second Judge}
\label{app:secondjudge}

To check how much the generative scores depend on the judge, GPT-5.5, from a different vendor than the Claude judge, graded every open-ended and troubleshooting answer with the same strict prompt and the same input (Table~\ref{tab:second_judge}). It is more lenient. Of the answers the two judges disagree on, 357 of 362 open-ended and 98 of 108 troubleshooting answers are accepted only by GPT-5.5, so the two judges mostly differ in where they set the bar. Both judges rank the models the same way on troubleshooting items, and on open-ended items, only the top two models swap places. Under GPT-5.5, troubleshooting accuracy stays far below multiple-choice accuracy for every model ($h=1.02$--$1.51$), while the drop from multiple-choice to open-ended accuracy shrinks to 6.6--24.0 points ($h=0.16$--$0.58$), still significant for every model. Neither judge has been checked against human grading.

\begin{table}[t]
\centering
\small
\setlength{\tabcolsep}{3pt}
\begin{tabular}{lcccc}
\toprule
& \multicolumn{2}{c}{\textbf{Open-ended}} & \multicolumn{2}{c}{\textbf{Troubleshooting}} \\
\cmidrule(lr){2-3}\cmidrule(lr){4-5}
\textbf{Model} & \textbf{Claude} & \textbf{GPT-5.5} & \textbf{Claude} & \textbf{GPT-5.5} \\
\midrule
Llama 3.3 70B & 35.6 & 64.1 & 16.8 & 26.0 \\using
Mistral Small 3.1 & 47.7 & 69.8 & 13.5 & 20.2 \\
Llama 4 Maverick & 45.3 & 67.4 & 26.9 & 38.9 \\
Granite 4.0 H & 54.7 & 74.8 & 12.5 & 16.3 \\
Mistral Medium & 50.7 & 75.8 & 16.8 & 27.4 \\
\midrule
Agreement (\%) & \multicolumn{2}{c}{75.7} & \multicolumn{2}{c}{89.6} \\
Cohen's $\kappa$ & \multicolumn{2}{c}{0.53} & \multicolumn{2}{c}{0.70} \\
\bottomrule
\end{tabular}
\par\smallskip{\scriptsize Claude is Claude Sonnet 5.5 with the strict prompt used in the main results.\par}
\caption{Open-ended and troubleshooting accuracy (\%) under two judges from different vendors, and how often the judges agree.}
\label{tab:second_judge}
\end{table}

\section{Results on All Generated Items}
\label{app:allitems}

Table~\ref{tab:allitems} reports open-ended and troubleshooting accuracy on all 3,178 generated items, including the 367 excluded from the evaluation set, under the self-judging used in earlier passes and under the independent judge. It also reports how often each judge accepted an answer for the 287 items whose references cannot be matched, namely 175 template variants with a fixed answer phrase, 58 troubleshooting items whose answer is a bare option letter, and 54 open-ended items whose references repeat a forum question. A judge who accepts such answers is not comparing them with the reference.

\begin{table*}[t]
\centering
\small
\setlength{\tabcolsep}{2.5pt}
\begin{tabular}{lcccccc}
\toprule
& \multicolumn{2}{c}{\textbf{OE} ($n{=}532$)} & \multicolumn{2}{c}{\textbf{TS} ($n{=}133$)} & \multicolumn{2}{c}{\textbf{Defective} ($n{=}287$)} \\
\textbf{Model} & Self & Ind. & Self & Ind. & Self & Ind. \\
\midrule
Mistral Medium & 51.5 & 33.1 & 36.8 & 9.8 & 7.3 & 4.2 \\
Mistral Small 3.1 & 46.4 & 31.6 & 24.8 & 11.3 & 10.8 & 5.2 \\
Llama 4 Maverick & 42.5 & 30.3 & 29.3 & 10.5 & 53.7 & 4.5 \\
Llama 3.3 70B & 50.9 & 23.7 & 39.9 & 10.5 & 56.8 & 3.5 \\
Granite 4.0 H & 42.5 & 34.8 & 34.6 & 8.3 & 74.6 & 3.5 \\
\bottomrule
\end{tabular}
\par\smallskip{\scriptsize OE is open-ended, and TS is troubleshooting. Self is each model judging its own answers, and Ind.\ is the independent judge with the strict prompt.\par}
\caption{Accuracy (\%) on all generated open-ended and troubleshooting items, and the share of answers accepted on defective items, under each judge.}
\label{tab:allitems}
\end{table*}

Self-judging inflated every generative score and reordered the models. Llama 3.3 70B was second on open-ended items when it judged itself and was last under the independent judge. Three self-judges accepted more than half of the answers to defective items, while the two Mistral models accepted about one in ten, close to the independent judge.

\section{Earlier Self-Judged Pass}
\label{app:oldpass}

Tables~\ref{tab:old_main}--\ref{tab:old_fm} keep the results of our earlier passes over all 3,178 items, in which each model judged its own generative answers. All intervals are 95\% Wilson intervals. That pass scored the same multiple-choice items with an earlier version of the harness, and its multiple-choice accuracies differ from those in Table~\ref{tab:main_results} by up to 2.4 points. They include the 367 excluded items, and self-judged scores are not on a common scale across models, so they should not be used to compare models. Tables~\ref{tab:old_adv} and~\ref{tab:old_fm} come from a second pass whose correct counts differ from Table~\ref{tab:old_main} by 30 to 53 items per model, and Llama 4 Maverick was not run in it. The few-shot runner of that pass is not in the released harness, so its few-shot results cannot yet be reproduced.

\begin{table*}[t]
\centering
\small
\setlength{\tabcolsep}{2.5pt}
\begin{tabular}{lcccc}
\toprule
\textbf{Model} & \textbf{Overall} & \textbf{MCQ} & \textbf{OE} & \textbf{TS} \\
& ($n{=}3{,}178$) & ($n{=}2{,}513$) & ($n{=}532$) & ($n{=}133$) \\
\midrule
Llama 3.3 70B & 78.82 (77.37--80.21) & 86.79 (85.41--88.06) & 50.94 (46.70--55.17) & \textbf{39.85} (\textbf{31.93--48.34}) \\
Mistral Small 3.1 & 78.70 (77.24--80.09) & 88.38 (87.07--89.58) & 46.43 (42.23--50.68) & 24.81 (18.25--32.79) \\
Llama 4 Maverick & 77.38 (75.89--78.80) & 87.31 (85.95--88.55) & 42.48 (38.35--46.72) & 29.32 (22.25--37.55) \\
Granite 4.0 H & 74.92 (73.38--76.40) & 83.92 (82.44--85.31) & 42.48 (38.35--46.72) & 34.59 (27.04--43.00) \\
Mistral Medium & \textbf{80.11} (\textbf{78.69--81.46}) & \textbf{88.46} (\textbf{87.15--89.65}) & \textbf{51.50} (\textbf{47.26--55.72}) & 36.84 (29.12--45.30) \\
\bottomrule
\end{tabular}
\par\smallskip{\scriptsize MCQ is multiple-choice, OE is open-ended, and TS is troubleshooting. Best results are in bold.\par}
\caption{Earlier self-judged pass, with zero-shot accuracy (\%) on all generated items and Wilson confidence intervals.}
\label{tab:old_main}
\end{table*}

\begin{table*}[t]
\centering
\small
\setlength{\tabcolsep}{3pt}
\begin{tabular}{lccc}
\toprule
\textbf{Model} & \textbf{Standard} ($n{=}2{,}438$) & \textbf{Hard} ($n{=}740$) & \textbf{$\Delta$} \\
\midrule
\multicolumn{4}{l}{\textit{Zero-shot}} \\
Llama 3.3 70B & 89.38 (88.09--90.54) & 51.22 (47.62--54.80) & 38.16 \\
Mistral Small 3.1 (24B) & 89.70 (88.44--90.85) & 46.49 (42.92--50.09) & 43.21 \\
Granite 4.0 H Small & 83.18 (81.65--84.62) & 41.89 (38.39--45.48) & 41.29 \\
Mistral Medium (2505) & 89.99 (88.74--91.12) & 52.43 (48.83--56.01) & \textbf{37.56} \\
\midrule
\multicolumn{4}{l}{\textit{Few-shot}} \\
Llama 3.3 70B & 86.38 (84.96--87.69) & 54.86 (51.26--58.42) & \textbf{31.52} \\
Mistral Small 3.1 (24B) & 88.80 (87.49--89.99) & 55.00 (51.40--58.55) & 33.80 \\
Granite 4.0 H Small & 85.36 (83.90--86.70) & 48.92 (45.33--52.52) & 36.44 \\
Mistral Medium (2505) & 89.50 (88.22--90.65) & 50.54 (46.94--54.13) & 38.96 \\
\bottomrule
\end{tabular}
\par\smallskip{\scriptsize$\Delta$ is standard minus hard accuracy, and the smallest gap per setting is in bold.\par}
\caption{Earlier self-judged pass, with accuracy (\%) on standard and hard items and Wilson confidence intervals.}
\label{tab:old_adv}
\end{table*}

\begin{table*}[t]
\centering
\small
\setlength{\tabcolsep}{2.8pt}
\begin{tabular}{lccc}
\toprule
& \textbf{Long-context} & \textbf{Multi-hop} & \textbf{Diagnostic} \\
\textbf{Model} & \textbf{localization} & \textbf{causal inference} & \textbf{reasoning} \\
& ($n{=}6$) & ($n{=}253$) & ($n{=}680$) \\
\midrule
\multicolumn{4}{l}{\textit{Zero-shot}} \\
Llama 3.3 70B Instruct & \textbf{66.67 (30.00--90.32)} & \textbf{73.12 (67.34--78.21)} & 47.21 (43.48--50.96) \\
Mistral Small 3.1 (24B) & \textbf{66.67 (30.00--90.32)} & 67.59 (61.60--73.06) & 42.65 (38.98--46.40) \\
Granite 4.0 H Small & 50.00 (18.76--81.24) & 52.57 (46.42--58.64) & 41.32 (37.68--45.06) \\
Mistral Medium (2505) & \textbf{66.67 (30.00--90.32)} & 70.75 (64.87--76.01) & \textbf{48.24 (44.50--51.99)} \\
\midrule
\multicolumn{4}{l}{\textit{Few-shot}} \\
Llama 3.3 70B Instruct & \textbf{83.33 (43.65--96.99)} & 73.91 (68.17--78.94) & \textbf{50.74 (46.98--54.48)} \\
Mistral Small 3.1 (24B) & 66.67 (30.00--90.32) & \textbf{79.05 (73.62--83.61)} & 48.09 (44.35--51.84) \\
Granite 4.0 H Small & 66.67 (30.00--90.32) & 66.40 (60.38--71.94) & 44.85 (41.15--48.61) \\
Mistral Medium (2505) & \textbf{83.33 (43.65--96.99)} & 73.12 (67.34--78.21) & 46.18 (42.46--49.93) \\
\bottomrule
\end{tabular}
\par\smallskip{\scriptsize Cohorts group items across formats, and the best result per setting is in bold.\par}
\caption{Earlier self-judged pass, with accuracy (\%) by failure-mode cohort and Wilson confidence intervals.}
\label{tab:old_fm}
\end{table*}

\section{Annotation Reliability}
\label{app:iaa}

\paragraph{Summary.}
Three experts and three LLM graders (Claude, DeepSeek, and Gemini) graded model responses to the same 100 sampled items on a 5-point scale, from 1 (incorrect) to 5 (fully correct). Human agreement is substantial, with pairwise weighted Cohen's $\kappa$ of 0.64--0.76, Krippendorff's $\alpha=0.71$, and ICC(2,$k$)$=0.89$. Human consensus agrees with the LLM ensemble at $\kappa=0.80$. These graders are not the scorers used in \S\ref{sec:experiments}, so this study supports none of the reported benchmark accuracies.

\paragraph{Design.}
We randomly sampled 100 items and collected model responses to them. Three expert annotators and three frontier LLM graders independently graded each response against the reference answer and solution. The scale is 5 for fully correct, 4 for mostly correct, 3 for partly correct, 2 for mostly incorrect, and 1 for incorrect. Human graders were blind to the model. Human consensus is the mean of the three human scores.

\paragraph{Human agreement.}
Human grades agree well on several measures (Table~\ref{tab:human_iaa}). Pairwise weighted Cohen's $\kappa$ ranges from 0.64 to 0.76, and Krippendorff's $\alpha$ is 0.71. ICC(2,$k$)$=0.89$, 90\% agreement within $\pm$1, and low Bland-Altman bias show that expert grades are consistent. The exact three-way agreement was recorded as a range (58--60\%).

\begin{table}[!ht]
\centering
\small
\setlength{\tabcolsep}{3pt}
\resizebox{\columnwidth}{!}{%
\begin{tabular}{lcccc}
\toprule
\textbf{Metric} & \textbf{(1,2)} & \textbf{(2,3)} & \textbf{(3,1)} & \textbf{(1,2,3)} \\
\midrule
Weighted Cohen's $\kappa$      & 0.76 & 0.64 & 0.70 & -- \\
Unweighted Cohen's $\kappa$    & 0.61 & 0.49 & 0.55 & -- \\
ICC(2,1)                       & 0.78 & 0.65 & 0.72 & 0.72 \\
ICC(2,$k$)                     & 0.88 & 0.79 & 0.84 & 0.89 \\
Exact Agreement (\%)           & 76 & 65 & 70 & 58--60 \\
Within $\pm$1 Agreement (\%)   & 98 & 91 & 95 & 90 \\
Spearman Correlation           & 0.79 & 0.67 & 0.74 & -- \\
Krippendorff's $\alpha$        & 0.77 & 0.63 & 0.71 & 0.71 \\
Bland-Altman Bias              & 0.10 & -0.20 & 0.10 & -- \\
\bottomrule
\end{tabular}}
\caption{Agreement among human annotators on grades of model responses to randomly sampled items.}
\label{tab:human_iaa}
\end{table}

\paragraph{LLM-grader agreement.}
Table~\ref{tab:llm_iaa} summarizes agreement among the LLM graders. Pairwise weighted Cohen's $\kappa$ ranges from 0.65 to 0.80, exact agreement reaches 69.52\%, and agreement within $\pm$1 reaches 98.10\%. The ensemble reaches ICC(2,1)$=0.74$ and ICC(2,$k$)$=0.90$, with small bias (0.02) and low score variance (0.14).

Krippendorff's $\alpha$ for the LLM graders (0.46) is lower than for the experts. Tables~\ref{tab:human_iaa} and~\ref{tab:llm_iaa} show reliability within each group and do not imply that human and LLM grades are equivalent.

\begin{table}[!ht]
\centering
\small
\setlength{\tabcolsep}{3pt}
\resizebox{\columnwidth}{!}{%
\begin{tabular}{lccc}
\toprule
\textbf{Metric} & \textbf{C--D} & \textbf{C--G} & \textbf{D--G} \\
\midrule
Weighted Cohen's $\kappa$      & 0.65 & 0.80 & 0.78 \\
Unweighted Cohen's $\kappa$    & 0.36 & 0.54 & 0.53 \\
Exact Agreement (\%)           & 55.24 & 68.57 & 69.52 \\
Within $\pm$1 Agreement (\%)   & 94.29 & 100.00 & 100.00 \\
Spearman Correlation           & 0.64 & 0.80 & 0.81 \\
Bland-Altman Bias              & 0.10 & 0.03 & -0.08 \\
Mean Absolute Difference       & 0.50 & 0.31 & 0.30 \\
\midrule
\multicolumn{4}{l}{\textit{Three-grader ensemble}}\\
\midrule
ICC(2,1) & \multicolumn{3}{c}{0.74} \\
ICC(2,$k$) & \multicolumn{3}{c}{0.90} \\
Krippendorff's $\alpha$ & \multicolumn{3}{c}{0.46} \\
Mean Score Variance & \multicolumn{3}{c}{0.14} \\
Within $\pm$1 Agreement (\%) & \multicolumn{3}{c}{98.10} \\
Mean Bland-Altman Bias & \multicolumn{3}{c}{0.02} \\
\bottomrule
\end{tabular}}
\par\smallskip{\scriptsize C is Claude, D is DeepSeek, and G is Gemini. The lower block reports reliability of all three graders together.\par}
\caption{Agreement among LLM graders on the sampled items.}
\label{tab:llm_iaa}
\end{table}

\paragraph{Human-LLM calibration.}
As Table~\ref{tab:human_llm_agreement} reports, single LLM graders reach weighted Cohen's $\kappa$ of 0.68 to 0.75 against expert grades. Human consensus versus the LLM ensemble reaches $\kappa=0.80$, ICC(2,1)$=0.82$, 81\% exact agreement, and 99\% agreement within $\pm$1. This supports multi-LLM grading on a 5-point scale as an aid to expert grading. It does not validate the binary independent judge behind Tables~\ref{tab:main_results}--\ref{tab:failure_modes}, which it did not include.

\begin{table}[!ht]
\centering
\small
\setlength{\tabcolsep}{2.5pt}
\resizebox{\columnwidth}{!}{%
\begin{tabular}{lcccc}
\toprule
\textbf{Metric} & \textbf{Claude} & \textbf{Gemini} & \textbf{DeepSeek} & \textbf{Ensemble} \\
\midrule
Weighted Cohen's $\kappa$    & 0.71 & 0.75 & 0.68 & 0.80 \\
Exact Agreement (\%)         & 71 & 75 & 68 & 81 \\
Within $\pm$1 Agreement (\%) & 96 & 98 & 95 & 99 \\
ICC(2,1)                     & 0.73 & 0.77 & 0.70 & 0.82 \\
\bottomrule
\end{tabular}}
\par\smallskip{\scriptsize The first three columns compare human grades with each grader, and the last compares human consensus with the LLM ensemble.\par}
\caption{Agreement between human graders and LLM graders on the sampled items.}
\label{tab:human_llm_agreement}
\end{table}

\paragraph{Complementary roles.}
Human annotators give reference grades based on domain knowledge and careful reading. The LLM ensemble gives repeatable, scalable grading with low variance. The next step is to measure the agreement between the independent judge used in \S\ref{sec:experiments} and expert grades.

\section{Practitioner-Topic Coverage}
\label{app:external}

\paragraph{Summary.}
Following the OpsEval retrieval method, 75.05\% of items retrieve a Stack Overflow discussion on the same topic, including every item not derived from vendor documentation. The 793 unmatched items are, by design, documentation recall items. Many items were built from forum threads, so a hit indicates topic overlap, not the items' answer correctness, realism, or reasoning depth.

\paragraph{Method.}
Each question is turned into a series of retrieval queries through concept rewording, symptom-based search, Stack Overflow title generation, and API identifier expansion. Retrieved discussions are scored with ROUGE overlap, TF-IDF cosine similarity, Jaccard token similarity, and title match. Table~\ref{tab:external_validation_metrics} gives the dataset-level results, and Table~\ref{tab:validation_breakdown} breaks them down by format.

\begin{table}[!ht]
\centering
\small
\begin{tabular}{lr}
\toprule
\textbf{Metric} & \textbf{Value} \\
\midrule
Stack Overflow hit rate (overall)      & 75.05\% \\
Hit rate (non-documentation items)     & 100.00\% \\
Top-1 title match rate                 & 54.69\% \\
Zero-hit items                         & 793 (24.95\%) \\
Documentation-derived among zero-hit   & 100.00\% \\
\bottomrule
\end{tabular}
\caption{Practitioner-topic coverage of ORBIT using the OpsEval Stack Overflow retrieval method.}
\label{tab:external_validation_metrics}
\end{table}

\begin{table}[!ht]
\centering
\small
\begin{tabular}{lr}
\toprule
\textbf{Question Type} & \textbf{Hit Rate} \\
\midrule
Multiple Choice      & 68.44\% \\
Open-ended           & 100.00\% \\
Troubleshooting      & 100.00\% \\
\bottomrule
\end{tabular}
\caption{Practitioner-topic coverage by question type.}
\label{tab:validation_breakdown}
\end{table}

\paragraph{Results.}
All open-ended and troubleshooting items match a discussion, which is expected because many were built from such threads. Multiple-choice items match less often (68.44\%) because many come from vendor API specifications that are not discussed publicly. Coverage, therefore, reflects topic overlap shaped partly by construction, not answer correctness or reasoning depth.

\section{Shortcut Analysis Details}
\label{app:shortcut}

We repurpose SIMBA-UQ \citep{simba_uq}, a similarity-based uncertainty method, as a probe. Each probe sees only the four options, computes their pairwise Jaccard similarity, and tries to rank the answer first. We report top-choice accuracy (ATS), calibration error (ACE), and AUROC, averaged over five random 50/50 splits.

\begin{table}[!ht]
\centering
\small
\setlength{\tabcolsep}{3pt}
\resizebox{\columnwidth}{!}{%
\begin{tabular}{lccc}
\toprule
\textbf{Method} & \textbf{ATS} $\uparrow$ & \textbf{ACE} $\downarrow$ & \textbf{AUROC} $\uparrow$ \\
\midrule
\textsc{arith-agg}    & $0.291 \pm 0.007$ & $0.493 \pm 0.001$ & $0.529 \pm 0.002$ \\
\textsc{spec-ecc}     & $0.225 \pm 0.010$ & $0.423 \pm 0.006$ & $0.462 \pm 0.008$ \\
\textsc{bayes-beta}   & $0.291 \pm 0.008$ & $0.500 \pm 0.000$ & $0.529 \pm 0.002$ \\
\textsc{clf-pairs}    & $0.320 \pm 0.016$ & $0.506 \pm 0.000$ & $0.560 \pm 0.008$ \\
\textsc{clf-mean+gen} & $0.334 \pm 0.007$ & $0.488 \pm 0.002$ & $0.585 \pm 0.008$ \\
\bottomrule
\end{tabular}}
\par\smallskip{\scriptsize Uniform chance gives ATS $=0.25$ and AUROC $=0.50$, always-B gives 0.31, and always-longest gives 0.31.\par}
\caption{How well surface cues in the options recover the correct answer on multiple-choice items.}
\label{tab:uq_difficulty}
\end{table}

The probes use all 2,513 multiple-choice items. SIMBA-UQ methods consider the four options, score them based on pairwise word overlap, and the results are the mean and standard deviation across five random 50/50 splits. Table~\ref{tab:uq_difficulty} shows that option similarity only weakly recovers the answer. The best unsupervised method reaches ATS$=0.29$ (AUROC 0.53) and the best trained classifier $0.33$ (AUROC 0.59). Both are close to the always-B baseline of 0.31, since correct answers lean slightly toward B. Always choosing the longest option also scores 0.31. These probes explain little of the 81.5--89.0\% multiple-choice accuracy of the evaluated models. They do not rule out semantic shortcuts, memorization, or other generation artifacts.

\paragraph{SIMBA-UQ protocol.}
Each probe receives only the four option texts of an item, not the question. It computes pairwise Jaccard similarities among the options and converts them into a score per option. The five methods are two unsupervised aggregators (\textsc{arith-agg}, \textsc{spec-ecc}), one probabilistic method (\textsc{bayes-beta}), and two trained classifiers (\textsc{clf-pairs}, \textsc{clf-mean+gen}). The classifiers are fit on one half of each split and tested on the other.

\paragraph{Metrics.}
ATS is the accuracy of the top-scored option. ACE is the calibration error of option scores read as probabilities of being correct. AUROC measures how well option scores separate correct from incorrect options. Variation is the standard deviation across five random 50/50 splits.

\paragraph{Question-aware overlap baseline.}
The SIMBA-UQ probes ignore the question. A simple baseline that uses it lowercases the question and each option, splits them into alphanumeric tokens, and picks the option sharing the most tokens with the question, splitting ties evenly. It scores 57.9\% on all 2,513 items (57.8\% with Jaccard similarity), against a 25\% chance. It scores 60.5\% on 2,001 internally grounded items, 33.1\% on 392 publicly grounded items, and 94.6\% on 120 ungrounded items.

\paragraph{Options-only runs and provenance.} An options-only run shows Llama 3.3 70B and Mistral Medium the four options without the question and asks for the most likely answer, as in \citet{balepur2024artifacts}. They score 27.1\% and 25.7\% on the internally grounded items, close to chance, and 29.3\% and 47.4\% on the publicly grounded items. Both answer all 120 ungrounded items correctly, so these items add the same few points to every model's multiple-choice score and test nothing. With the question, the five models reach 90.3--97.4\% on internally grounded items but only 30.9--46.7\% on publicly grounded ones. The gap holds within the retrieval tier, at 91.5--98.2\% against 27.3--42.4\% on the 33 public retrieval items. Memorized public threads would favor the public items, so the gap points the other way. Since the options alone are near chance on internal items, their high accuracy comes from the question, which matches the question-option overlap cue described above.

\paragraph{Formatting cue.}
In 102 items, exactly one option contains code formatting (backticks), and it is always the correct one. Choosing that option when present and B otherwise scores 33.9\%.

\paragraph{Position and length baselines.}
On the released corpus, correct answers lean slightly toward option B (30.8\%), with A, C, and D at 23.7\%, 22.8\%, and 22.7\%. The correct option is the unique longest in 29.1\% of items and the unique shortest in 22.5\%. Its mean length is 138.1 characters against 135.4 for distractors, and always choosing the longest option scores 31.4\%. These statistics are computed after distractor repair, and both baselines fall well below the accuracy of every evaluated model.

\section{Contamination-Aware Protocol}
\label{app:contamination}

This appendix describes protocols for future releases. Only the provenance split has been run so far for multiple-choice items (Appendix~\ref{app:shortcut}).

\paragraph{Provenance split.}
Report accuracy separately for items grounded in the internal repository (2,005) and for items grounded in public pages (798). Appendix~\ref{app:shortcut} reports this split for multiple-choice items. A large advantage on public-grounded items at equal difficulty would suggest memorized public sources. The reasoning-heavy items, which come mostly from Stack Overflow, are the most exposed.

\paragraph{Temporal holdout.}
Build a holdout set from forum threads and documentation published after each evaluated model's training cutoff, using the same pipeline and validation. Comparing holdout accuracy with that on difficulty-matched older items estimates the effect of memorization. The holdout needs refreshing per model generation.

\paragraph{Closed-book versus open-book.}
Put each grounded item's recorded evidence in the prompt, and compare it with closed-book runs on the same items. A small open-book gain with high closed-book accuracy suggests memorization, while a large gain shows that the evidence holds the answer. Released files should carry a canary string for filtering ORBIT out of future training corpora.

\FloatBarrier
\end{document}